\documentclass[Preprint,12pt,authoryear]{elsarticle}

\usepackage{subfigure}
\usepackage{setspace}
\usepackage{geometry} 
\usepackage{framed,multirow}
\usepackage{float}
\usepackage{latexsym}
\usepackage{bm}
\usepackage{amsmath}
\usepackage{booktabs}
\usepackage{xcolor}
\usepackage{array}
\usepackage{algorithmic}
\usepackage{dblfloatfix}
\usepackage{verbatim}
\usepackage{mathrsfs}
\usepackage{amsthm,amsmath,amssymb}
\usepackage{lineno,hyperref}
\usepackage{geometry}
\journal{}

\usepackage{numcompress}

\begin{document}

\begin{frontmatter}
\title{GraNet: Global Relation-aware Attentional Network for ALS Point Cloud Classification}


\author{Rong Huang, Yusheng Xu, Uwe Stilla}

\address{Photogrammetry and Remote Sensing, Technical University of Munich (TUM), Munich, Germany\\
}
\begin{abstract}
Semantic labeling is an essential but challenging task when interpreting point clouds of 3D scenes. 
As a core step for interpretation, semantic labeling is to annotate every point in the point cloud with a label of semantic meaning, which plays a significant role in many point cloud related applications. 
Especially for airborne laser scanning (ALS) point clouds, precise annotations can considerably broaden its use in various applications.
However, accurate and efficient semantic labeling is still a challenging task, due to the sensor noise, complex object structures, incomplete points, and uneven point densities.
In this work, we propose a novel neural network focusing on semantic labeling of ALS point clouds, which investigates the importance of long-range spatial and channel-wise relations and is termed as global relation-aware attentional network (GraNet). 
GraNet first learns local geometric description and local dependencies using a local spatial discrepancy attention convolution module (LoSDA). 
In LoSDA, the orientation information, spatial distribution, and elevation differences are fully considered by stacking several local spatial geometric learning modules and the local dependencies are embedded by using an attention pooling module. 
Then, a global relation-aware attention module (GRA), consisting of a spatial relation-aware attention module (SRA) and a channel relation aware attention module (CRA), are investigated to further learn the global spatial and channel-wise relationship between any spatial positions and feature vectors. 
The aforementioned two important modules are embedded in the multi-scale network architecture to further consider scale changes in large urban areas.
We conducted comprehensive experiments on two ALS point cloud datasets to evaluate the performance of our proposed framework.
The results show that our method can achieve higher classification accuracy compared with other commonly used advanced classification methods.
The overall accuracy ($OA$) of our method on the ISPRS benchmark dataset can be improved to 84.5 {\%} to classify nine semantic classes, with an average $F_1$ measure ($AvgF_1$) of 73.5 {\%}, which outperforms other strategies.
Besides, experiments were conducted using a new ALS point cloud dataset covering highly dense urban areas.
\end{abstract}


\end{frontmatter}


\section{Introduction}

{A}{irborne} laser scanning (ALS), as one of the most important systems using the light detection and ranging (LiDAR) technique. 
By carrying LiDAR devices on an aircraft or UAV, ALS has advantages of quickly acquiring large-scale and high-precision ground information \citep{vosselman2017contextual,li2019improving}, which enable it to be utilized in a wide variety of applications such as 3D city modeling \citep{moussa2010automatic, lafarge2012creating,yang2013automated}, land cover and land use mapping \citep{yan2015urban}, forestry monitoring \citep{ reitberger20093d, polewski2015detection}, construction monitoring \citep{bosche2015value,xu2018a, huang2020temporal}, change detection \citep{hebel2013change}, powerline inspection \citep{clode2005classification,guo2015classification}, and deformation monitoring \citep{alba2006structural,olsen2010terrestrial}. However, acquired ALS point clouds usually provide points with 3D coordinates and attributes (e.g., intensities, incident angles, or numbers of returns), but without semantic information indicating labels of the scanned ground objects \citep{huang2020deep}, which hinders further applications like urban mapping or building reconstruction. 

To parse semantic information of the 3D scene from ALS point clouds, one practical solution is semantic labeling. 
The primary goal of the semantic labeling for ALS point clouds is to annotate every point in the point cloud with a label of semantic meaning, in accordance with geometric or radiometric information provided by the point itself and its neighborhood. 
This can be achieved via a classification of acquired points. 
For traditional supervised classification methods using handcrafted features, their performance relies on two crucial factors: distinctive features and discriminative classifiers. 
Features of a point are an abstracted mathematical expression that can quantifiably describe the characteristics of a point, indicating the point belonging to a certain category. 
In previous work, to create discriminative features,  studies have exploited both point geometry and inherent attributes. 
Successful cases include contextual features from spatial distributions and directions of points \citep{yang2017computing}, eigenvalue based features from covariance matrix of point coordinates \citep{chehata2009airborne, weinmann2015semantic, weinmann2015distinctive}, waveform-based features from transformation \citep{jutzi2010investigations,zhang2011full}, 2D projected patterns \citep{zhao2018classifying}, elevation values and height differences \citep{maas1999potential,gorgens2017method,sun2018classification}, and orientations of points from normal vectors \citep{rabbani2006segmentation}.
However, designing good handcrafted features is a critical and difficult task, which requires a good understanding of the scanned objects and highly depends on empirical tests \citep{xu2019classification}.
Classifier refers to a mathematical function or transformation implemented by algorithms or strategies, which projects input features to a category. 
A well-designed classifier should maximum the discrimination of features of various semantics. 
Regarding point cloud classification, a considerable amount of classifiers have been introduced and tested,  including AdaBoost \citep{chan2008evaluation}, support vector machines (SVM) \citep{mallet2011relevance}, composite kernel SVM \citep{ghamisi2017}, random forest (RF) \citep{chehata2009airborne}, Hough forest \citep{yu2016bag}, and conditional random fields (CRF) \citep{niemeyer2014contextual, weinmann2015contextual, yao2017semantic, vosselman2017contextual,li2019higher}. 
For supervised classification, a classifier is trained using the generated features and the corresponding labels, so that the parameters of the classifier can be optimized for inferring categories of points from input features. 
Then, the trained classifier can be used to predict labels of points from other test areas. 
\begin{figure*}[h!]
	\begin{center}
		\includegraphics[width=1\textwidth]{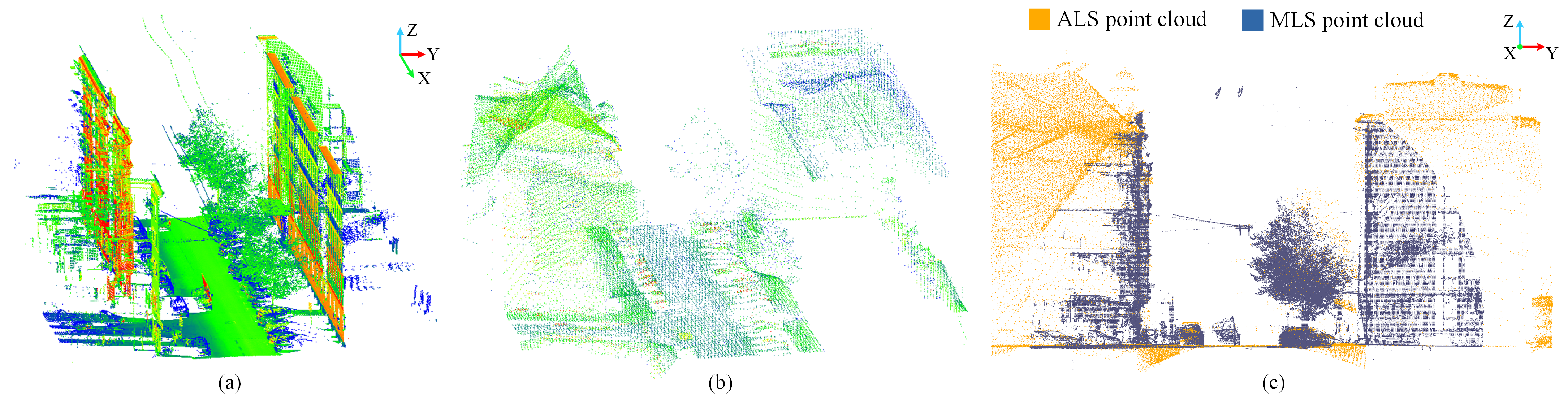}
		\caption{Difference between (a) MLS and (b) ALS point clouds of a same street scene. (c) Different points distribution of two point clouds.}
		\label{fig:alsandmls}
    \end{center}
\end{figure*}

Compared with classification approaches using handcrafted features, classification methods using learned features can automatically discover the feature representations needed for classification from raw points, requiring less prior knowledge, and avoiding sophisticated feature design. 
Feature learning is usually achieved via dictionary learning or deep learning. 
Especially, deep-learning-based methods for point cloud classification are getting increasingly popular recently, which provides end-to-end solutions that significantly mitigate burdens in feature design and processing complexity. 
For a neural network-based method, its layer structure and parameters can implicitly express the spatial interactions between 3D points, facilitating the feature representations. 
Tennumerous neural networks have achieved remarkable performance in a wide range of applications. 
Wellknown examples include VoxNet \citep{maturana2015voxnet}, MultiViewCNN \citep{su2015multi}, PointNet \citep{qi2017pointnet}, PointNet++ \citep{qi2017pointnet++}, PointCNN \citep{li2018pointcnn}, PointSIFT \citep{jiang2018pointsift}, SPG \citep{landrieu2018large}, RandLA-Net \citep{hu2020randla}, and many more. 
However, these methods were mainly proposed as solutions for indoor and close-range applications in the fields of computer vision, autonomous driving, and robotic.
When it comes to ALS point clouds acquired in large-scale urban scenarios, additional aspects should be considered when using neural network-based methods. 
This is because ALS point clouds have some unique characteristics comparing with 3D point clouds acquired from LiDAR systems on other platforms \citep{li2020geometry}, for example, significant variations of object scales, discrepancy distributions of elevations, consistent orientations of objects. In Fig.~\ref{fig:alsandmls}, we provide an illustration comparing MLS and ALS point clouds of the same street scene. 
As seen from the figure, we can find that spatial distributions of points in MLS and ALS point cloud are totally different. 
Points of the former one concentrate in the vertical direction and the road surface, while points of the latter are evenly located in horizontal directions but with significant elevation differences.
Thus, for approaches of ALS point cloud classification, scale factor, local elevation discrepancy, and object orientations should always be taken into consideration.  
In some recent work \citep{li2020geometry,li2020dance,wen2020directionally}, these factors have been considered.
However, all these factors represent merely local geometric characteristics of points. First, in this case, the receptive fields are limited if only local convolutional operations are applied. Second, geometric or radiometric characteristics of points from different categories may be similar in the local scope but can be distinguished by considering points in long range but with more similar characteristics. Thus, exploiting long-range relations may provide additional and rich information provided in large receptive fields. Inspired by many recent advances in relation-based and attention-based neural networks widely used in various 2D applications, such as person re-identification \citep{zhang2020relation}, image classification \citep{wang2017residual}, and object detection \citep{hu2018relation,xu2019spatial}, long-range dependencies can be further investigated in the task of point cloud classification. 

In light of this, we propose a global relation-aware attentional network (GraNet) for point feature embedding in ALS point cloud classification, counting three aspects: 
(i) introducing a local spatial encoding module for 3D point clouds consider spatial discrepancy (i.e., orientation information on the horizontal plane and elevation differences in the vertical direction) of each point and preserving the local correlations between points by considering attentions of local structure; 
(ii) proposing a global relation-aware attention module (GRA), including a spatial relation-aware attention module (SRA) and a channel relation-aware attention module (CRA), to further investigate the long-term dependencies between any spatial positions and feature channels; 
(iii) embedding the convolution and attention modules into a multi-scale framework. 
Major contributions of our work are summed up as follows:
\begin{itemize}
\item A novel local spatial discrepancy attention convolution module (LoSDA) considering horizontal orientation and elevation differences of 3D points is designed, and local point relationships are further considered by adding an attention pooling module.
\item A GRA module, consisting of two submodules: SRA and CRA, are exploited to express the importance of long-range dependencies between 3D points, especially for point clouds in large-scale areas.
\item A multi-scale network architecture, which embeds the LoSDA module and the GRA module, is presented for ALS point cloud classification. 
\end{itemize}

The remainder of this paper is organized as follows:
Section \ref{Sec:RW} discusses the related work. 
Section \ref{Sec:MT} describes the GraNet for ALS point cloud classification, including the LoSDA module, the GRA module, and the network architecture. 
Section \ref{Sec:EX} describes the datasets and the evaluation metric. 
Section \ref{Sec:ER} displays the experimental results. 
Then, Section \ref{Sec:DIS} presents the ablation study and discusses the complexity and efficiency of the network. 
Section \ref{Sec:CO} provides the conclusions and future directions for our work.

\section{Related work}\label{Sec:RW}

Generally, deep learning methods for point cloud classification can be categorized into four major types, namely projection-based methods, voxel-based methods, point-based methods, and graph-based methods.

\subsection{Projection-based methods}

The core idea of projection-based methods is to project points from the 3D Euclidean space to 2D planes or manifold space so that the projected data can utilize CNN approaches designed for 2D data.
Values (e.g., grayscales, intensities, or densities) of the projected data will represent either geometry (e.g., elevations) or attributes (e.g., RGB colors) of the original 3D points. 
The most commonly used 2D projected representation is imagery.
The 2D rendered image derived from virtual cameras of different viewing positions \citep{su2015multi} is an example. 
Through the use of multiple 2D rendered images, the 3D geometry of an object can be delineated. 
The rendered 2D images of an object are then applied to a 2D CNN network for object classification. 
According to reported studies, projection-based methods have demonstrated success in various classification applications using large scale LiDAR point clouds. 
In \cite{yang2017convolutional}, geometric features and full-waveform attributes were generated from local neighboring points of each point and assigned with $x-$ and $y-$ coordinates to be a pixel in a 2D image.
The generated 2D images encapsulated either geometric and radiometric information and were then fed into 2D CNNs.
With the predicted labels of 2D pixels and a back-projection,  the labeling of 3D points could be achieved. 
In \cite{yang2018segmentation},  based on the previous strategy of \citep{yang2017convolutional}, a multi-scale CNN and 2D images of features extracted from neighborhoods of various scales were developed, which obtained a better performance of classification. 
In \cite{boulch2018snapnet}, snapshot images with pixels, in which RGB colors and depths were encoded, were applied to a 2D fully convolutional network. 
Once the labels of 2D pixels were obtained, they were back-projected to the original 3D space to eventually fulfill 3D point classification. 
In \cite{zhao2018classifying}, 2D images of multi-scale contextual information, including features of height, intensity, and roughness, were attained to represent the original 3D LiDAR points. 
Then, a multi-scale CNN was applied to conduct a classification of 3D points using these 2D images.
Apart from 2D images, a digital surface model (DSM), as 2.5D data, is also adopted to represent 3D points of ALS data, since points in ALS data always reveal an even distribution in horizontal directions and lack of vertical distribution. 
In \cite{chen2017multi}, the DSM generated from 3D points was utilized as one input, which was fed into a two-stream deep neural network. 
Generally speaking,  the design of the neural network based on the projection method is directly inherited from the existing 2D CNN solutions, and there is almost no need to adjust the network structure. However, these methods are deficient in presenting information from the depth direction and inevitably cause errors in rendering and interpolation.

\subsection{Voxel-based methods}

Voxel-based approaches structure the 3D space into regular voxel grids and project discrete 3D points into these voxels.  
Then, 3D points will be represented by the spatial occupancy of voxels so that a 3D convolution with a cubic template can be applied.
Voxnet \citep{maturana2015voxnet} is one early example, which directly transformed points to 3D voxels assigned with occupancy and implemented a 3D CNN to predict class labels of 3D objects. 
In \cite{engelcke2017vote3deep}, as an improvement, point clouds were voxelized into grid structures, then the voting procedure was introduced in a 3D CNNs.
Similar to grey values of 2D pixels, occupancy are the most commonly used attributes that could be assigned to 3D voxels. However, there are also some other strategies to represent points with voxel. 
For instance, in \cite{wang2015voting}, values of voxels were encoded by attributes generated from spatial positions of all points within this voxel. 
Besides, irregular-shaped 3D grid structures (e.g., voxels may have different sizes or cuboid shaps) can also be used for organizing 3D points. 
For example, the octree structure was introduced to CNNs in \cite{wang2017cnn}, wherein normal vectors of points in each leaf node were averaged as as voxel values and then fed into CNN.
Similarly, Kd-trees were also utilized to structure discrete 3D points \citep{klokov2017escape}.
Voxel structures can also combine with pixels. For example, in \cite{qin2019semantic}, both voxels and pixels were used as inputs in the proposed VPNet for semantic labeling of ALS data. 
The generation of voxels provided contextual information from the local area.
On the contrary, auxiliary structures can be utilized to assist voxels as well. 
For instance, in \cite{zhou2018voxelnet}, unified features of every voxel were extracted via an additional feature encoding layer on the basis of region proposal network. 
These features would tackle the sparsity of 3D points. 
In \cite{qi2016volumetric}, two aforementioned schemes (i.e., voxel-based convolution and feature encoding layers) were combined into a multi-orientation volumetric CNN. 
The features of each orientation were generated with a shared network, and results of an image-based CNN were also integrated. 
However, similar to the shortcomings of projection-based methods, the voxelization process, either regular-shaped or irregular shaped ones, definitely lead to a loss of spatial information since this is a sampling process. In this case, aliasing is inevitable due to the setting of the voxel resolution. 
Moreover, points of different categories may be rasterized into a voxel with the same label, which adds the ambiguity and decreases the accuracy.
The 3D structure requires considerably larger memory consumption and computational cost than 2D images, which hinders applying and developing this type of method.

\subsection{Point-based methods}

Point-based methods use discrete points as input to networks.
As a milestone, the emergence of PointNet \citep{qi2017pointnet} started the trend of directly using discrete points in deep neural networks. 
PointNet and its variants \citep{qi2017pointnet++, qi2018frustum} showed remarkable performance on popular benchmarks with either indoor \citep{armeni20163d, dai2017scannet} or outdoor \citep{geiger2012we, hackel2017isprs,zhang20193} applications. 
One of the key innovations of PointNet is to consider the unstructured and disordered characteristics of 3D points through a transformation and align points into the same orientation frame. 
Thus, it can establish an end-to-end framework in which 3D points can be classified without preprocessing.
Moreover, for each point, local and global features were considered and learned in PointNet.
In \cite{yousefhussien2018multi}, the multi-scale frame was developed to embed PointNet, for achieving a classification of large-scale ALS point clouds. 
In \cite{li2018pointcnn}, PointCNN with a new network structure was proposed to learn an $X$ transformation of points to deal with permutations and centralization. 
In \cite{jiang2018pointsift}, PointSIFT encoded 3D information of point orientations, then the module was embedded in the multi-scale frame of PointNet++. 
In \cite{li2020geometry}, a geometry-attentional network was designed, which improved PointSIFT by embedding dense hierarchical structure and elevation-attention module. 
In this work, low-level geometric vectors were introduced to induce the learning of high-level local pattern representation, which increased the discrimination of geometric awareness of features. 
In \cite{li2020dance}, a density-aware convolution module was introduced to directly work on 3D point sets and deal with uneven density distribution of 3D point clouds. 
Additionally, a context encoding module was designed to regularize the global semantic context. 
In \cite{wen2020directionally},  a directionally constrained fully convolutional neural network utilized a novel directionally constrained point convolution module to encode local context in an orientation-aware way by considering projected 2D receptive fields.
Moreover, point-based methods are also used as an encoder for extracting deep features, which can be integrated with other optimization algorithms. 
For instance, in \cite{huang2020deep}, PointNet++ with hierarchical data augmentation was proposed to learn deep features of points and then optimized by a manifold-based feature embedding. 

\subsection{Graph-based methods}

Rather than directly using discrete points as input, points, as well as the contexts, can be structured by a graph. 
A graphical model could naturally represent the spatial space of 3D scenes \citep{landrieu2018large, xu2019classification}.
The graph-structured data is then fed into a newly designed network. GraphCNN is an encouraging instance, which has shown promising results on different applications \citep{simonovsky2017dynamic,landrieu2018large, wang2018dynamic}. 
In the graph-structured data, the edges between the points are created for generating the topology of the graph. 
Recently, the attention mechanism is becoming increasingly popular, as it can provide scores of importance for parameters. 

Compared with the traditional classification methods, these deep learning-based methods generate less noise in the classification results. 
As a by-product of the meta-parameterization of the network \citep{landrieu2017structured}, it has regularity. 
However, this spatial regularity is uncontrollable. 
At the same time, relatively speaking, for deep learning techniques (such as PointNet), the classification result depends on the point sampling and block cutting methods in the preprocessing and the interpolation method in the postprocessing \citep{huang2020deep}. 
In the process of splitting, downsampling, and interpolation, especially at the boundary of the split point set, artifacts and noise will be generated. 
Thus, additional constraints are necessary to be introduced as input for a network.

\section{Methodology} \label{Sec:MT}

Our proposed network includes three major parts, namely one LoSDA module and the GRA module consisting of SRA and CRA, which are all embedded in a multi-scale network architecture. 
For the LoSDA module, we embed the spatial distribution encoding (SDE), directional feature encoding (DFE), and evaluation difference encoding (EDE) in the local point embedding and builds the local correlation in the local neighborhood by using an attention pooling module. 
In the GRA module, the contextual information is further investigated under the global scope. 
For the multi-scale network architecture, we develope an encoder-decoder frame on the basis of PointNet++ \citep{qi2017pointnet++} architecture.   

\subsection{Local spatial discrepancy attention convolution module}

\begin{figure*}[h!]
	\begin{center}
		\includegraphics[width=1\textwidth]{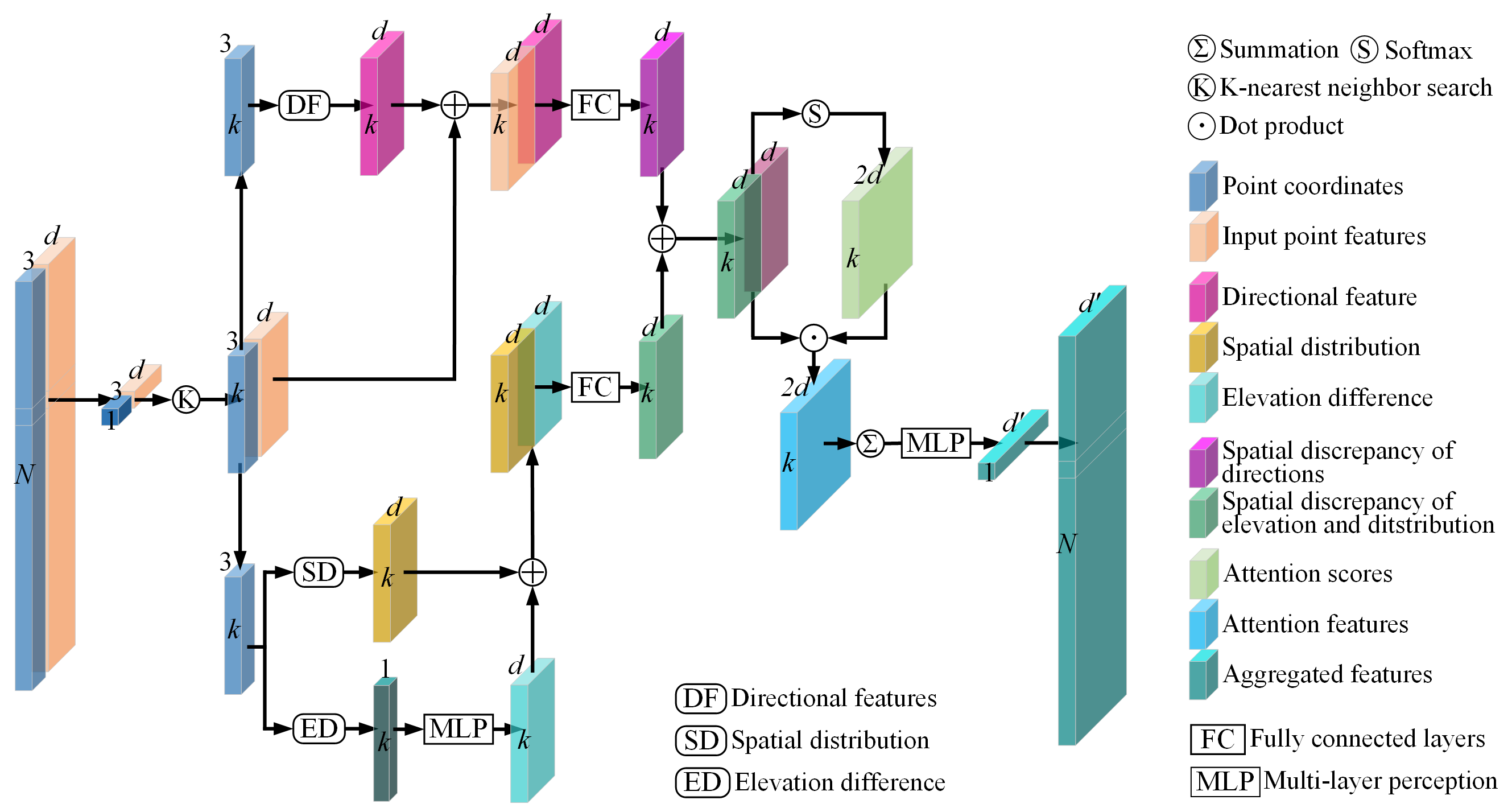}
		\caption{Exploited local spatial attention encoding module.}
		\label{fig:lsam_mod}
    \end{center}
\end{figure*}

The LoSDA module consists of SDE, DFE, and EDE, and an attention pooling module, which sequentially implements the description of local geometric characteristics and considers the local dependencies.
In Fig.~\ref{fig:lsam_mod}, we illustrate the scheme of the LoSDA module we used, which is applied to each 3D point and outputs aggregated features. 
In this module, encoding the spatial discrepancy for a point is conducted based on the local spatial information provided by $k$-nearest neighboring (KNN) points. 
For the definition of KNN points of a point, we employe the local feature aggregation strategy proposed in \cite{hu2020randla}. 
To be specific, for an input point $p_i$ together with attributes or intermediate learned features $f_i$, the local spatial encoding will explicitly embed local spatial information of all its KNN points. 
The metric of KNN is Euclidean distances. 
Based on the KNN points, we can encapsulate the spatial discrepancy of a point from the local context, which considers the spatial distribution, directional features, and elevation differences of points. 
By encoding information from these three aspects, we can get an explicit representation of the local spatial information for a given point.

\subsubsection{Spatial distribution encoding}

As stated in \cite{hu2020randla}, given the input point $p_i$ and its KNN point set $P_i = \{ p_i^{1}, p_i^{2},p_i^{3},...,p_i^{K}\}$, the use of KNN points can make point features be aware of their relative spatial locations, indicating the local spatial distribution of points. 
To encode the spatial distribution of an input point $p_i$ in its KNN point set $P_i$, their relative point positions will be utilized, which is calculated as follows:
\begin{equation}
    r_i^{j} = MLP(p_i \oplus p_i^{j} \oplus (p_i - p_i^{j}) \oplus ||p_i - p_i^{j}||),
\end{equation}
where $\oplus$ is the concatenation operation, and $|| \cdot ||$ calculates the Euclidean distance between the neighboring and input points. 

\subsubsection{Directional feature encoding}

The directional feature of a point can also be induced from the spatial locations of points in the local area, which describes the orientation information in the horizontal directions. 
To encode the directional feature of the input point $p_i$, we utilize the point-wise local feature descriptor proposed by \citep{jiang2018pointsift}, which depicts directional information of eight orientations. 
Specifically, we implement a selection of eight nearest neighbors from the KNN point set $P_i$ in each of the eight octants by ordering three coordinates. 
Since distant points provide little information for a description of local patterns, when no point exists in point set $P_i$, we duplicate $p_0$ as the nearest neighbor of itself.
We further process features of those neighbors that reside in a $2 \times 2 \times 2$ cube for local pattern description centering
at $p_0$. 
Many previous works ignore the structure of data and do max pooling on feature vectors along $d$ dimensions to get new features. However, we believe that ordered operators such as convolution can better exploit the structure of data. 
Thus we propose orientation-encoding convolution, which is a three-stage operator that convolves the $2 \times 2 \times 2$ cube along $X$, $Y$, and $Z$ axis successively. 
Formally, the features of neighboring points is a vector $V$ of shape $2 \times 2 \times 2 \times d$, where the first three dimensions correspond to three axes. Slices of vector $M$ are feature vectors, for example $M_{1,1,1}$ represents the feature from top-frontright octant.
The three-stage convolution is formulated as:
\begin{equation}
        \begin{aligned}
        V_{x}   &= g(Conv(Wx, V ))  \in R_{1 \times 2 \times 2 \times d}\\
        V_{xy}  &= g(Conv(Wy, V_{x}))  \in R_{1 \times 1 \times 2 \times d}\\
        V_{xyz} &= g(Conv(Wz, V_{xy})) \in R_{1 \times 1 \times 1 \times d}
        \end{aligned},
\end{equation}
where $W_x \in R_{1 \times 2 \times 2 \times d}$,  $W_y \in R_{1 \times 2 \times 1 \times d}$, and $W_z \in R_{1 \times 1 \times 1 \times d}$ are weights of convolution operator (bias is omitted for clarity). 
In this work, we set $g( \cdot)$ = $ReLU( \cdot )$. 
Finally, we will get a $d$ dimension feature by reshaping $V_{xyz} \in R_{1 \times 2 \times 2 \times d}$, integrating information from eight spatial orientations and obtains a representation that encodes orientation information.

\subsubsection{Elevation difference encoding}

The elevation difference of points is calculated to delineate the positioning discrepancy in the vertical direction. 
To encode elevation difference of the input point $p_i$, the coordinates in $z$- direction of the local neighboring point set of $p_i$ is extracted and formed a 1-dimensional vector, and then the feature vector is embedded to the same d-dimensional feature using a MPL layer. The MLP layer is formed by a 1 $\times$ convolution layer, a batch normalization layer, and a ReLU activation layer. Thus, the elevation difference encoding can be formulated as:
\begin{equation}
    z_i^{j} = MLP(z_i),
\end{equation}
where $z_i$ is the $z$ coordinates of the input KNN point set $P_i$.

\subsubsection{Attention pooling}

In the attentive pooling, as stated in \cite{hu2020randla}, it will aggregate the obtained point features $\hat{F}_i$ of $p_i$. 
As shown in Fig.~\ref{fig:lsam_mod}, the attention scores are achieved by a shared MLP and a followed softmax, which is defined as follows:
\begin{equation}
    s_i^{j} = CAS(\hat{f}_i^{j}, W),
\end{equation}
where $W$ stands for the learnable weights of a shared MLP. Then, these features are weighted by the attention scores and summed as follows:
\begin{equation}
    \Tilde{f}_i = \sum_{i=1}^{K} (\hat{f}_i^{j} \cdot s_i^{j}),
\end{equation}
where $\Tilde{f}_i $ stands for the output aggregated feature of the input point $p_i$.

\subsection{Global relation-aware attention module}

Inspired by the work in \cite{zhang2020relation}, we design the GRA module, which directly works on 3D points to investigate the long-range relations between points and high-dimensional feature vectors. 
The GRA module consists of two sub-modules: SRA and CRA.

\subsubsection{Spatial relation-aware attention module}

\begin{figure*}[h!]
	\begin{center}
		\includegraphics[width=0.85\textwidth]{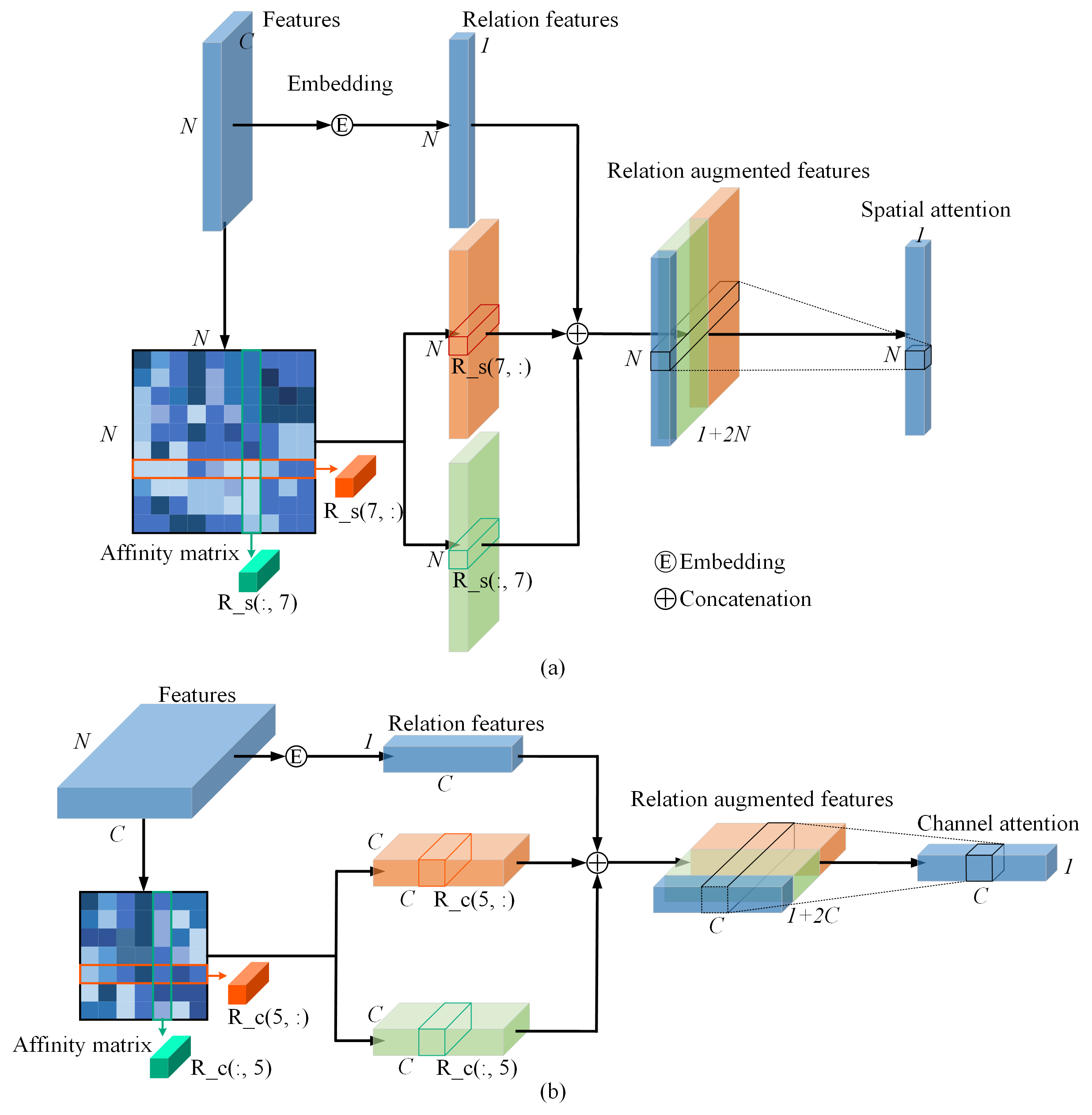}
		\caption{Illustration of the GRA module. (a) is the SRA module. (b) is the CRA module.}
		\label{fig:gra_module}
    \end{center}
\end{figure*}

In order to capture the long-range dependencies between points, the SRA module is applied to 3D points with intermediate deep features. 
In the former local attention module (i.e., attention pooling module in LoSDA), convolution operations are conducted with small receptive fields on feature maps. 
However, in order to learn the importance of a feature node, features should be compared under the global scope.  
Thus, we introduce the SRA module, which consists of three major steps, namely the calculation of relations (named as affinity matrix) between all nodes, the formation of relation-augmented features, and the calculation of attention scores for each single node. The details will be explained in the following texts, as illustrated in Fig.~\ref{fig:gra_module}.

Given a feature tensor $\textbf{X} \in \mathcal{R}^{N\times C}$ presenting a point set with $N$ points and $C$ channels from an immediate layer in the network. 
We regard the $C$-dimensional feature vector of each spatial point as a node in the graph.
Then, all the feature nodes representing the spatial positions can form a graph $G_s$ with $N$ nodes. 
The pairwise relation from node $n_i$ to node $n_j$ can be calculated using a dot product, which is termed as affinity:
\begin{equation}
    a_{i,j} = \alpha_s(x_i)^T \beta_s(x_j),
\end{equation}
where $\alpha_s$ and $\beta_s$ are two embedding functions implemented by a MLP layer. Then, the bidirectional relations between node $n_i$ and $n_j$ are described by the pairwise affinity value and the relations between all the points can be presented by an affinity matrix $\textbf{A}_s \in \mathcal{R}^{N \times N}$.

As illustrated in Fig.~\ref{fig:gra_module}, relation features are generated by extracting the column and the row at the spatial position from the affinity matrix. 
When calculating the attention  scores of the $n_i$ feature node, the original feature vector from the point itself should also be considered to further utilize both the non-local information and its relation to the local original information. 
Thus, the spatial relation-augmented features can be denoted as:
\begin{equation}
    y_i^s = [pool(\alpha_s(x_i)),\beta_s(r_i)],
\end{equation}
in which $\alpha_s$ and $\beta_s$ denote embedding functions for the original features and the global relation features. $\alpha_s$ denotes a MLP layer and the embedded features are further fed into a pooling layer after the embedding operation using $\alpha_s$. Here a max pooling is utilized. $\beta_s$ is also made of a MLP layer.
Since these two types of features do not come from the same domain, they are embedded separately and then concatenated to obtain spatial relation-augmented features.

The global relation features contain rich contextual information. 
In order to further mine valuable information from them, we propose to infer attention via a learnable model. 
The spatial attention values can be obtained by:
\begin{equation}
    a_i^s = Sigmoid(ReLU(y_i^s)).
\end{equation}
Then, the attention scores of each nodes can formed a attention matrix $\textbf{a}_s$.
Finally, the spatial relation-aware features can be obtained by:
\begin{equation}
    \textbf{Y}_s = \textbf{a}_s*\textbf{X}.
\end{equation}

\subsubsection{Channel relation-aware attention module}

To further consider the dependencies between channels, as shown in Fig.~\ref{fig:gra_module}, we use a similar strategy and design a channel relation-aware attention module. The main difference is that the relations are calculated between channel-wise nodes instead of spatial positions.

Given an intermediate feature tensor $\textbf{X} \in \mathcal{R}^{N\times C}$ presenting a point set with N points and C channels. 
We treat the point set as a graph $G_c$ with $C$ nodes. 
Thus, the pairwise relation from node $c_i$ to node $c_j$ can be calculated as:
\begin{equation}
    r_{i,j} = \alpha_c(x_i)^T\beta_c(x_j),
\end{equation}
where $\alpha_c$ and $\beta_c$ are two embedding functions, which are achieved by a MLP layer. 
Then, the relations between all feature nodes can be represented by an affinity matrix $\textbf{A}_c \in \mathcal{R}^{C\times C}$. Subsequently, for each feature node, we concatenate the original feature vector with the pairwise relations between all the nodes and generate a new relation-augmented feature vector as:
\begin{equation}
    y_i^c = [pool(\alpha_c(x_i)), \beta_c(r_i)],
\end{equation}
where $\alpha_c$ and $\beta_c$ also denote the embedding functions implemented by a MLP layer, while the embedding function $\alpha_c$ is followed by a pooling layer.
The channel attention value can then be obtained using a similar procedure to the procedure conducted on the corresponding spatial module:
\begin{equation}
    a_i^c = Sigmoid(ReLU(y_i^c)),
\end{equation}
where $a_i^c$ is attention score for each node and forms the attention matrix $\textbf{a}_c$.
Finally, the channel relation-aware features can be obtained by multiplying attention scores:
\begin{equation}
    \textbf{Y}_c = \textbf{a}_c*\textbf{X}.
\end{equation}

\subsubsection{Integration of the GRA module}

\begin{figure*}[h!]
	\begin{center}
		\includegraphics[width=0.8\textwidth]{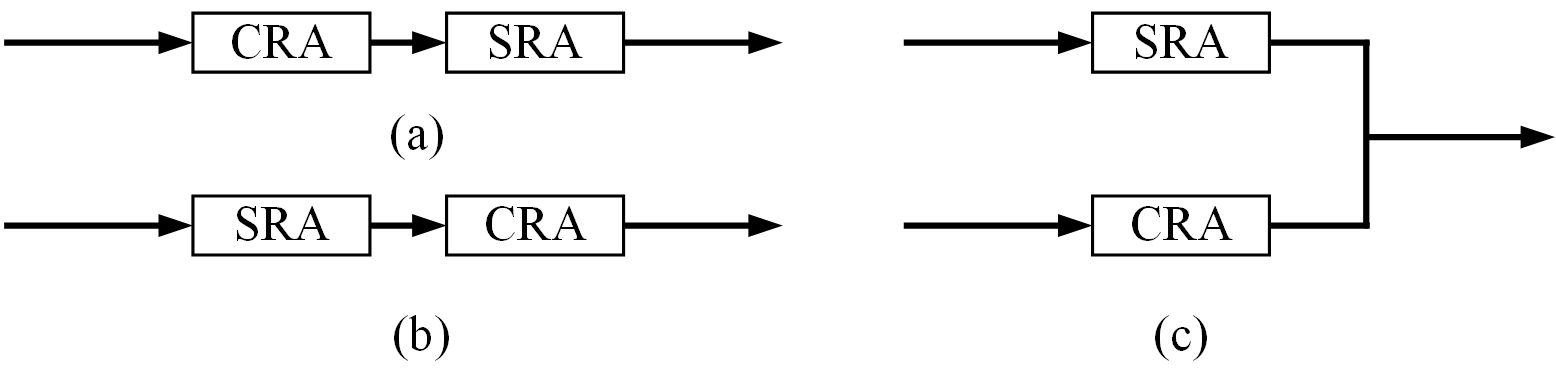}
		\caption{Illustration of the different configurations of the GRA module. (a) Mode 1. (b) Mode 2. (c) Mode 3.}
		\label{fig:gra_configuration}
    \end{center}
\end{figure*}

In order to make full use of the SRA and CRA modules, we further integrate these two modules. 
In this work, we investigate three patterns for the configuration. 
In the following text, to make the elaboration clearly and briefly, the configurations of SRA and CRA modules are terms as the GRA module. 
As illustrated in Fig.~\ref{fig:gra_configuration}, there are three modes in constructing the GRA module by combining the SRA and CRA modules. 
Mode 1 is a serial configuration, where the SRA module follows the CRA module. 
Mode 2 is also a serial configuration, but in this configuration, the CRA module comes after the SRA module. 
Mode 3 is a parallel pattern in which the intermediate deep features obtained after feeding to the CRA and SRA modules parallelly are concatenated to obtain the final output of the GRA module. 
The effect of different configurations will be tested in Section \ref{Sec:DIS}.

\subsection{Multi-scale network framework}

\begin{figure*}[h!]
	\begin{center}
		\includegraphics[width=\textwidth]{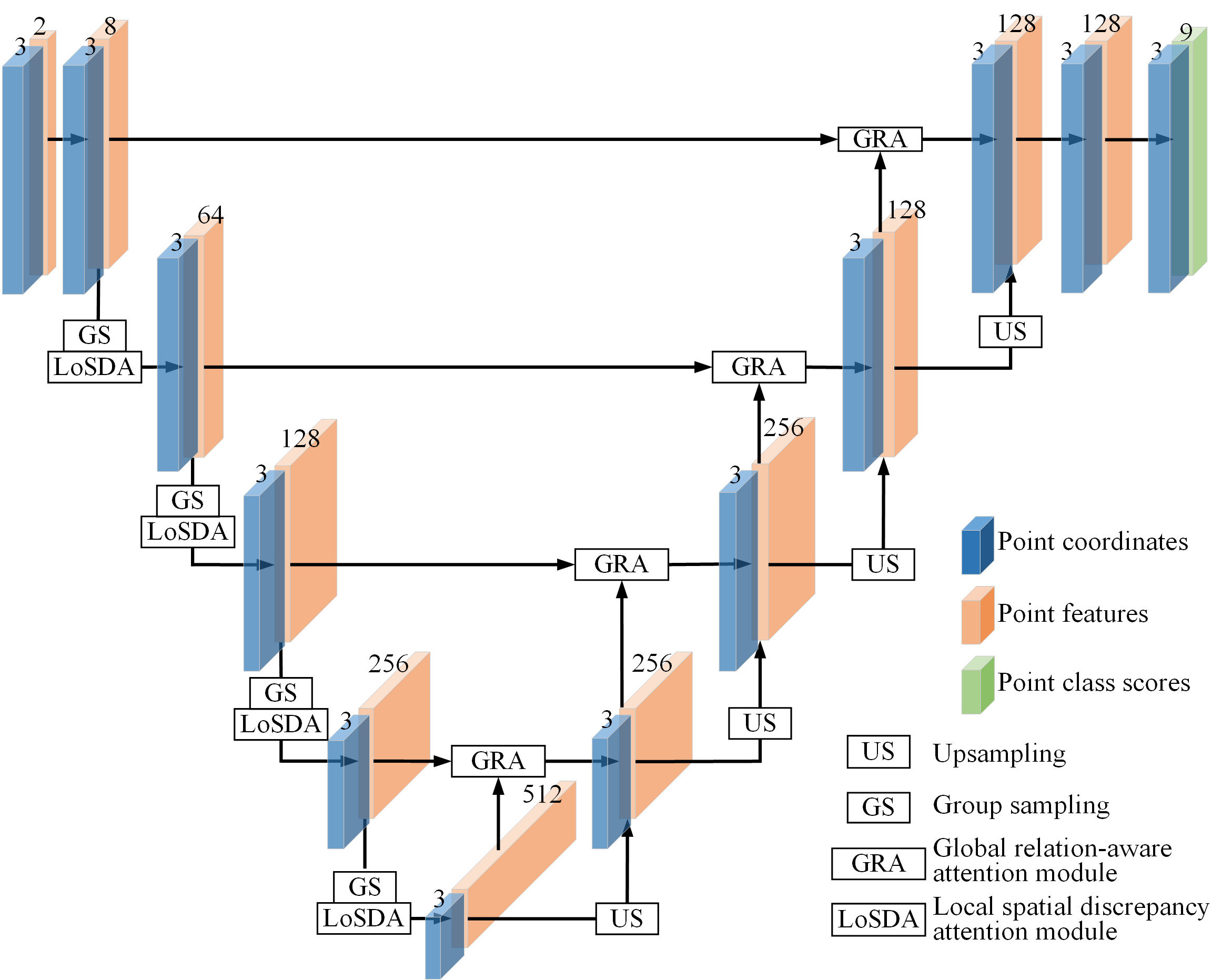}
		\caption{Illustration of the multi-scale network framework.}
		\label{fig:nn_structure}
    \end{center}
\end{figure*}

Our multi-scale network framework is inspired by the frame of PointNet++ \citep{qi2017pointnet++}. 
The LoSDA module and the GRA module are embedded in the multi-scale framework. 
In Fig.~\ref{fig:nn_structure}, we provide a detailed illustration of the network architecture. 
Here, the modules of grouping and sampling proposed in PointNet++ are integrated with the aforementioned LoSDA module. 
The grouping and sampling module take in $N \times d$ input standing for $N$ points and $d$ dimensional feature for each point. 
The output is $N' \times d'$ corresponding to $N'$ downsampled points having $d'$ dimensional feature. 
The downsampling is implemented by finding $N'$ centroids with farthest point sampling, assigning points to centroids, and then calculate the embedding of centroids by feeding features from the LoSDA module. 
The interpolation module is a distance-weighted linear interpolation, which upsamples the input points. 
The interpolation is fed with input points of size $N$, and outputs upsampled points of size $N'$, keeping feature dimensionality stable. 
The interpolation upsamples points and assigns them with the original features of the input points. Moreover, the features are concatenated with the features from the corresponding layer with the same number of points in the encoder. The fused features are then fed into the GRA module to generate the final output for features of the upampled points. 
The rule of assigning features to the new upsampled points is according to the $k$ nearest search of points by Euclidean distance in 3D space.

\subsubsection{Local spatial encoding}

Based on the multi-scale architecture, points are grouped and sampled to generated sub-pointsets on each group sampling layer. Based on the constructed neighborhood by the group sampling, instead of using the PointNet unit as the local spatial encoder, the local features are obtained by feeding the point itself and its local neighborhood to the LoSDA module.

\subsubsection{Relation-aware information fusion}

In the original multi-scale network structure proposed by PointNet++, high-level features are interpolated and then concatenated with low-level features directly in the feature propagation layer. 
The fused features are then fed into a shared MLP without further considering the different contextual information proposed different levels of features. 
To consider long-term dependencies between points, we propose a relation-aware information fusion module, namely the GRA module, to fuse high-level features and low-level features in the multi-scale network structure. 
As illustrated in Fig.~\ref{fig:nn_structure}, high-level features and low-level features are first concatenated to generate the fused features. 
Afterward, fused features are fed into the GRA module to obtain global augmented attentional information. 

\subsection{Details of the entire network architecture}

\begin{figure*}[h!]
	\begin{center}
		\includegraphics[width=\textwidth]{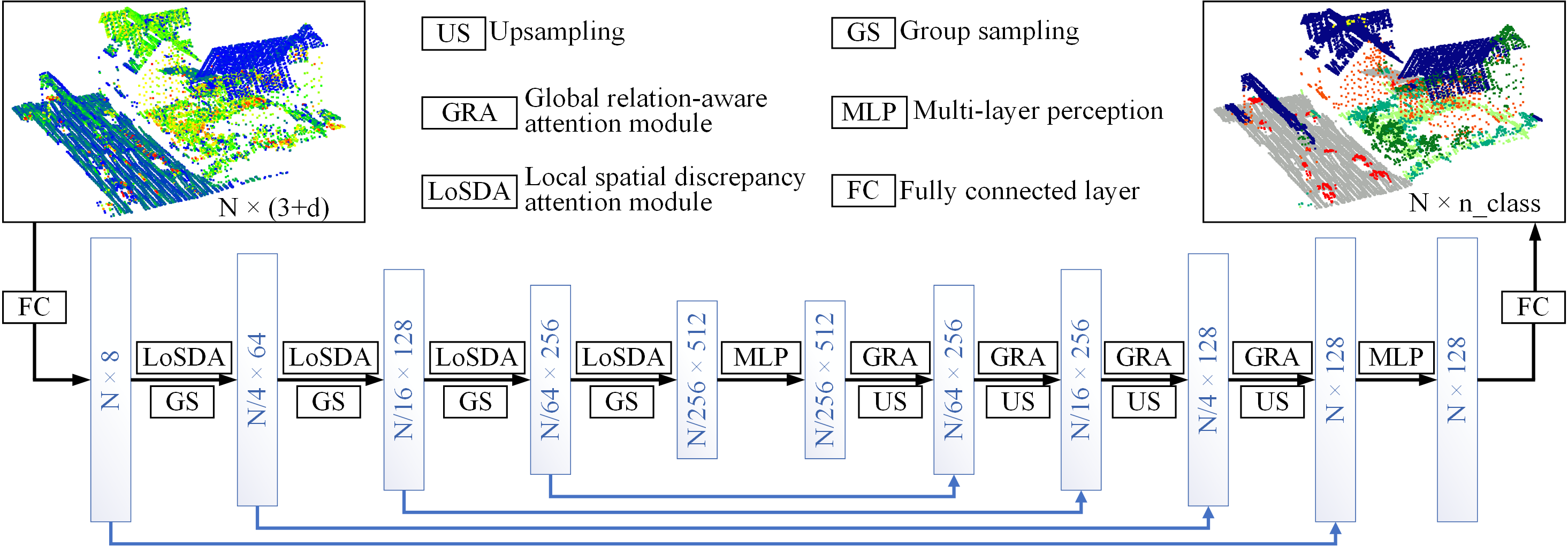}
		\caption{Detailed network architecture.}
		\label{fig:nn_details}
    \end{center}
\end{figure*}

Fig.~\ref{fig:nn_details} illustrates the detailed architecture of our network, which is implemented by a classic encoder-decoder architecture with skip connections. 
The input point cloud is initially fed to a fully connected layer to extract point features based on the original point features except for point coordinates. 
Three encoding and decoding layers are then used to learn features for each point. 
At last, a fully-connected layer is used to predict the semantic label of each point. 
The input is a large-scale original point cloud with a size of $N \times d$ (the batch dimension is dropped for simplicity), where N is the number of points, $d$ is the feature dimension of each input point. 
For our testing datasets, each point is represented by its 3D coordinates, intensity, and numbers of return. 
The encoder has three layers, progressively reducing the size of the point clouds and increasing the feature dimensions of each point. 
Each encoding layer consists of the LoSDA module and a grouping and sampling module. 
The point cloud is downsampled with a four-fold decimation ratio. 
In particular, only 25\% of the point features are retained after each layer. 
Meanwhile, the feature dimension of each point is gradually increased each layer to preserve more information. 
The decoder also has three layers that are used after the above encoding layers. 
For each layer in the decoder, we first use the KNN algorithm to find one nearest neighboring point for each query point. The point feature set is then upsampled through a nearest-neighbor interpolation. 
Next, the upsampled feature maps are concatenated with the intermediate feature maps produced by encoding layers through skip connections, after which the GRA module is applied to the concatenated feature maps. 
The semantic label of each point is predicted through a fully-connected layer.

\section{Experiments}\label{Sec:EX}
In this section, we present two ALS datasets, which were utilized to evaluate the performance of our method. The first dataset is the ISPRS benchmark dataset \citep{cramer2010dgpf,rottensteiner2012isprs}. The second one is the Large-scale Aerial LiDAR Dataset for Semantic Labeling in Dense Urban Areas (LASDU) \citep{ye2020large}, a large scale ALS dataset acquired from a highly-dense urban area. The evaluation metric for the performance of methods is also presented in this section.

\subsection{Testing datasets}
\begin{figure*}[h!]
	\begin{center}
		\includegraphics[width=1\textwidth]{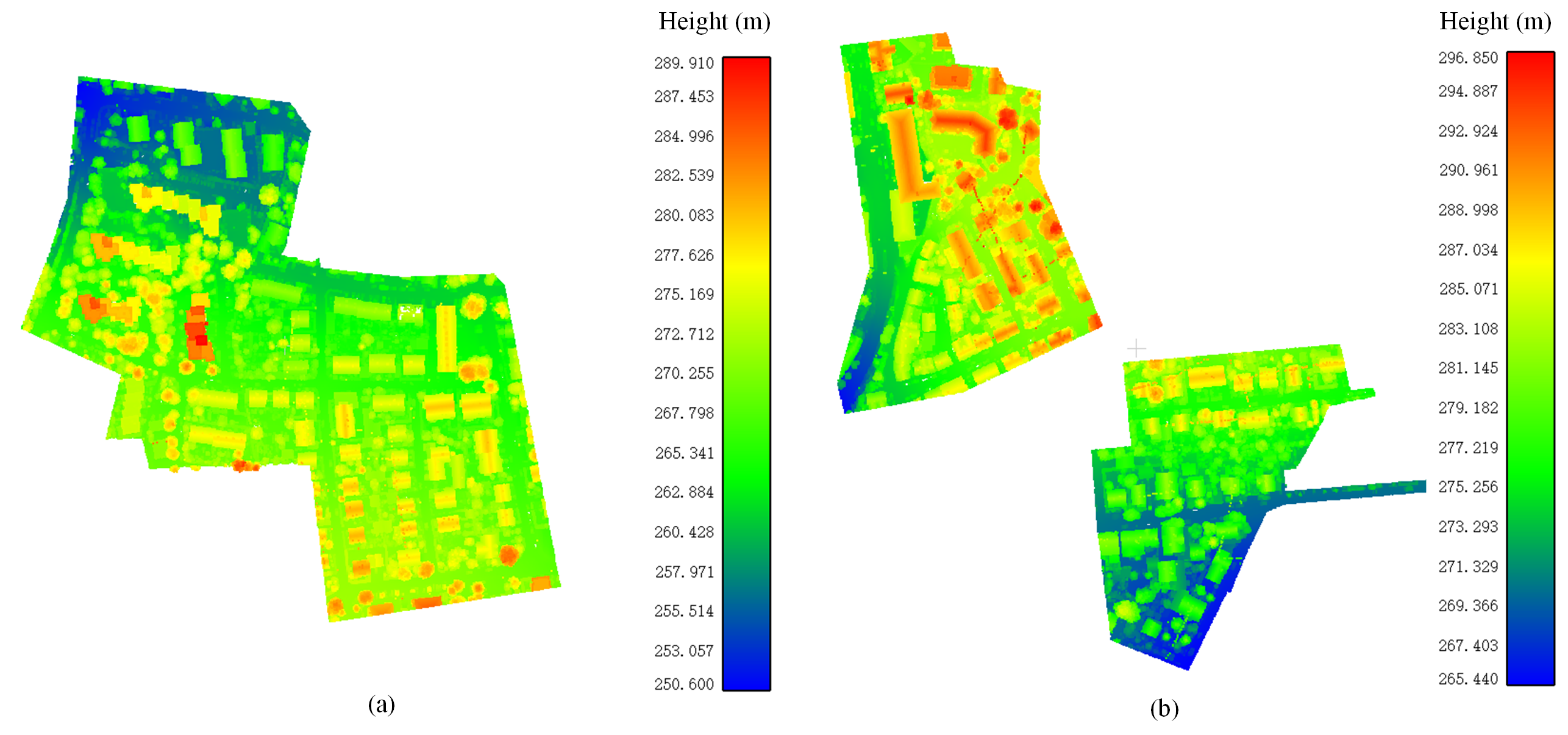}
		\caption{Top view of the ISPRS benchmark dataset colored by elevations. (a) Training and (b) test areas. It should be noted that the scales of two color bars are different.}
		\label{fig:area_ISPRS}
    \end{center}
\end{figure*}

The ISPRS benchmark dataset is an ALS benchmark dataset obtained in August 2008 using a Leica ALS50 system with an average flying height of 500m and a 45$^\circ$ field of view.
In this dataset, all points comprise 3D coordinate information and spectral information (i.e., intensity values and the number of returns). The covering area is located in the city center of Vaihingen, Germany, which is a highly dense area and involves many objects from different categories. This dataset's point density is about 4 points/$m^2$ and the numbers of points for training and testing are 0.75 million and 0.41 million, respectively. The data has been manually labeled to nine semantic categories, including powerline, low vegetation, impervious surfaces, cars, fences/hedges, roofs, facades, shrubs, and trees. The test and the training areas are illustrated in Fig.~\ref{fig:area_ISPRS}.

\begin{figure*}[h!]
	\begin{center}
		\includegraphics[width=1\textwidth]{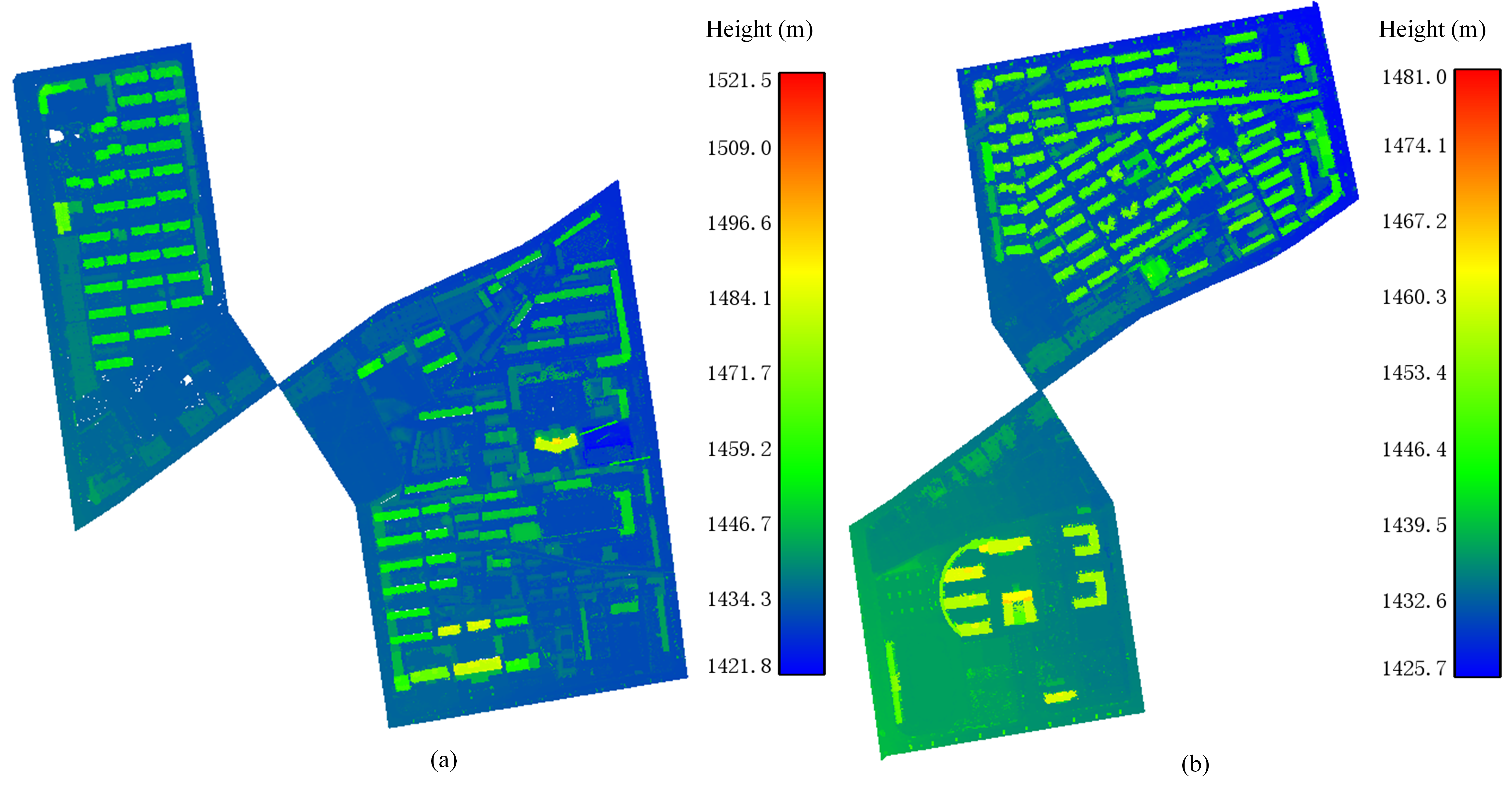}
		\caption{Top view of the LASDU dataset colored by elevations. (a) Training and (b) test areas. It should be noted that the scales of two color bars are different.}
		\label{fig:area_LASDU}
    \end{center}
\end{figure*}

The LASDU dataset is a part of data obtained in the campaigns from HiWATER (Heihe Watershed Allied Telemetry Experimental Research) project \citep{li2013heihe}. 
The study area is located in the valley along the Heihe River in the northwest of China. 
The covering study area is nearly flat, and the average elevation is about 1550 m. 
The dataset was acquired in July 2012 by a Leica ALS70 system onboard an aircraft with a flying height of about 1200 m. 
The average point density was approximately 3 points/$m^2$, and the vertical accuracy ranges between 5–30 cm. 
One part of the dataset was manually labeled, covering an urban area of around 1 $km^2$, with highly-dense residential and industrial buildings. 
The total number of annotated points is approximately 3.12 million. In the dataset, five categories of urban objects were considered: ground (e.g., artificial ground, roads, and bare land), buildings, trees (e.g., tall and low trees), low vegetation (e.g., bushes, grass, and flower beds), artifacts (e.g., walls, fences, light poles, vehicles, and other artificial objects). 
The entire labeled point cloud of the investigating area has been divided into four areas. The numbers of points in these four areas are around 0.77 million, 0.59 million, 1.13 million, and 0.62 million, respectively. 
In Fig.~\ref{fig:area_LASDU}, the training and testing datasets are shown, with the elevation of the study area given. 
From Fig.~\ref{fig:area_LASDU}, we can find that the annotated area is nearly flat, and the maximum difference of elevation is only around 70 m.

\subsection{Evaluation metric}

The following evaluation metrics are utilized to assess the performance of our method: precision (Pr), recall (Re), $F_1$ measure ($F_1$), overall accuracy (OA), and average $F_1$ measure($AvgF_1$). Among these metrics, the classification performance on every single category is evaluated using precision, recall, and $F_1$ measure, whereas $OA$ and $AvgF_1$ measures are applied to evaluate the performance of methods on the entire test data.

\section{Results} \label{Sec:ER}

\subsection{Results using ISPRS dataset}

\subsubsection{Implementation details}

\begin{figure*}[h!]
	\begin{center}
		\includegraphics[width=\textwidth]{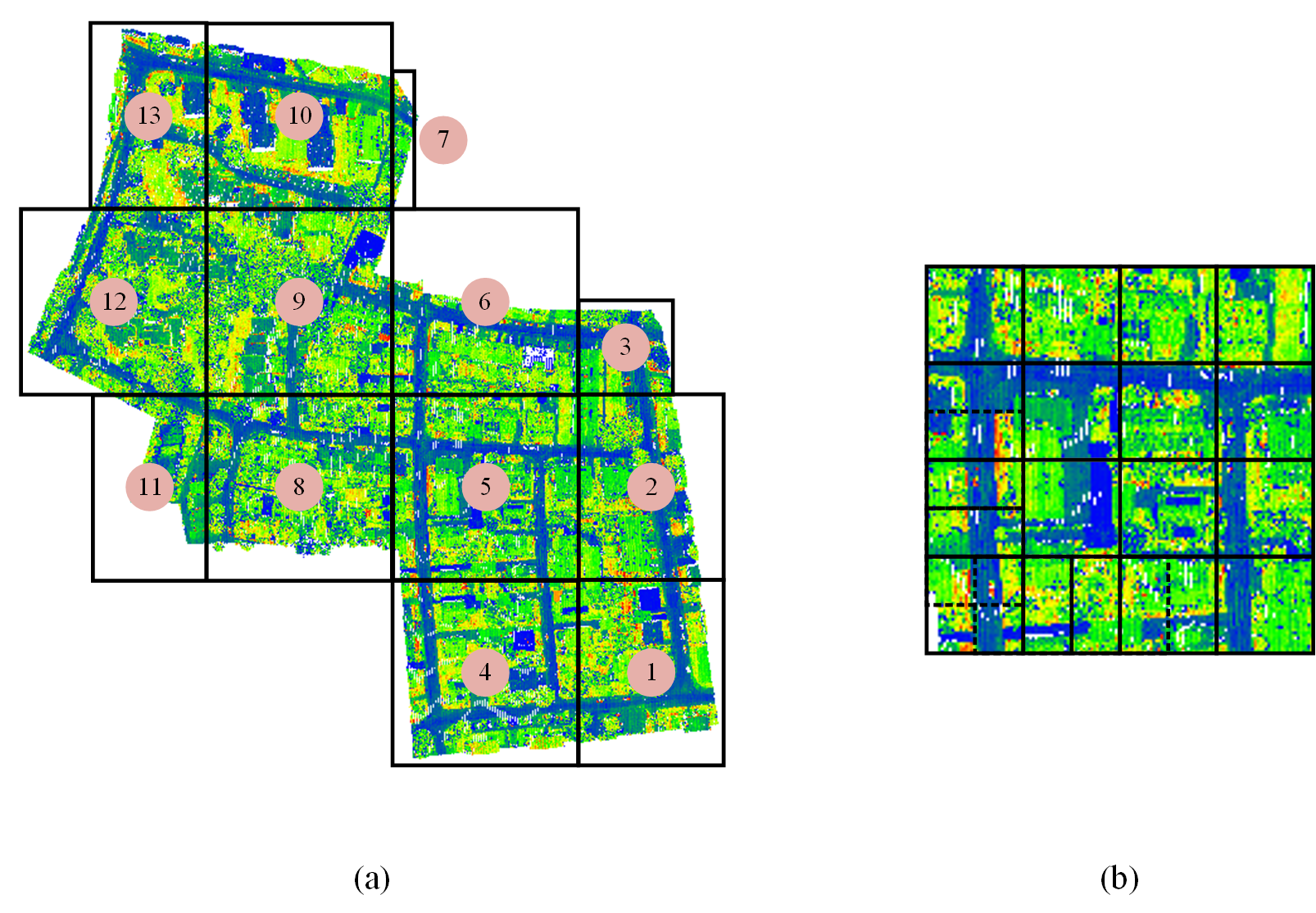}
		\caption{Illustration of the data division of the ISPRS benchmark dataset. (a) The division of the large training blocks. (b) The further subdivision of each training block.}
		\label{fig:isprs_split}
    \end{center}
\end{figure*}

In order to conduct the training process, we divided the dataset into several point blocks as illustrated in Fig.~\ref{fig:isprs_split}, similar to the strategy used in \cite{li2020geometry}. 
Considering the covering area of the dataset, the target size for each block in this step was set to be 100 m $\times$ 100 m. 
The training dataset was then divided into 13 non-overlapped blocks, which consisted of two blocks for validation and 11 blocks for training. Among these blocks, block 13 and block 4 were selected for validation considering the balance of different categories in the validation data. Each block was further divided into 25 m $\times$ 25 m blocks with a 12.5 m overlap in both $x-$ and $y-$directions. 
Besides, the test dataset was also processed using the same procedure for the training and validation datasets. 
After the two-step subdivision, the uneven densities of the dataset and each block hold points whose data amount varies from hundreds to ten thousand points. 
To meet the requirement for the input for the networks, we randomly sampled points for each block to generate sub-pointsets for training and test processes. 
A fixed value of 4096 was chosen as the size for each sub-pointset. Moreover, each point within the pointset was represented by a 5D vector considering the provided information that comprises the $x-$, $y-$, $z-$coordinates, intensities, and return numbers.

The training and testing of all our models were conducted in the framework of Tensorflow and performed on an NVIDIA TITAN X (Pascal) 12 GB GPU. 
In our models, when constructing a local neighborhood for each point, the number of neighboring points was all set as 32, the same as the hyperparameter in PointNet++. 
In the training process, we used Adam optimizer with an initial learning rate of 0.001, a momentum value of 0.9, and a batch size of 4. 
The learning rate was iteratively reduced based on the current epoch by the factor of 0.7. 
Moreover, the training process lasted for 1000 epochs totally, and the weights were saved if the loss decreased.

\subsubsection{Comparing with other PointNet-based methods}

Based on the obtained classification results, We compare our results with the other four methods, including PointNet \citep{qi2017pointnet}, PointNet++\citep{qi2017pointnet++}, Hierarchical Data Augmented PointNet++ (HDA-PointNet++) \citep{huang2019embedding}, and PointSIFT \citep{jiang2018pointsift}.
PointNet \citep{qi2017pointnet} is a neural network that directly uses discrete points as input, preserving the permutation invariance of unstructured points. 
PointNet proposed a unified framework for point clouds relating tasks ranging from object classification to semantic segmentation. 
PointNet++ \citep{qi2017pointnet++} is an improvement of PointNet, applying PointNet to extract features from local neighborhoods of points. 
A hierarchical architecture was designed to capture scale awarded features from the local context of points. 
HDA-PointNet++ is an improved method based on the original PointNet++, which was proposed in \cite{huang2020deep} that known as the multi-scale deep features (MDF) embedded in the hierarchical deep feature learning (HDL) method. 
This method proposed a hierarchical data augmentation approach, dividing point clouds into multi-scale pointsets for boosting the capability of dealing with scale variations of objects. 
The last one is PointSift \citep{jiang2018pointsift}, inspired by SIFT. It combined scale awareness and orientation-encoding the two fundamental properties to improve the 3D shape description ability.
Primarily, it designed an orientation-encoding unit to describe the information in different directions.

Table \ref{tab:result_pointnet} lists the classification results of the four aforementioned methods and our method, which are all under the framework of the multi-scale network structure. 
As shown by the table, our proposed neural network can outperform the other four reference PointNet++ based methods. In respect of the $OA$, we can find that our method can achieve the best results with an $OA$ of 84.5{\%}. 
First, it can be seen that PointNet shows lower ability when dealing with this kind of fine-grained situation, especially when considering the classification accuracy of some small objects, such as powerlines, cars, and fence hedges. 
Compare to the baseline method for our strategy, PointNet++, the $OA$ increases by 5.4{\%}. 
Additionally, compared with the improved solution, PointSIFT, our method also achieve better results with an increment of $OA$ by 1.8{\%} and $AvgF_1$ by 5.6{\%}. 
It indicates the effectiveness of our proposed local spatial encoding method and relation-aware strategy compared with purely encoding local neighborhood and orientation information with MLP. 
Furthermore, our method also achieves higher classification accuracy for most categories than the other PointNet++ based methods. 
It is also worth mentioning that HDA-PointNet++ shows the best results in some large-scale categories, such as low vegetation and impervious surface, compared with the other methods, although it does not provide competitive $OA$ and $AvgF_1$. 
The main reason is that for deep neural networks directly handling describe points, the subdivision and sampling of the entire point clouds in urban scenarios is a critical step that directly impacts the classification results. 
By the use of the multi-scale subdivision and sampling, the scale information can be considered, especially for the objects with larger-scale dependencies.

\begin{table*}[ht!]
    \centering
    \caption{Comparing with results of the ISPRS benchmark dataset using different PointNet++ based methods (Values in {\%}). Noted that the highest values in $OA$, $AvgF_1$, and $F_1$ for each category are marked with bold texts. \label{tab:result_pointnet}}
    \resizebox{\textwidth}{!}
    {\begin{tabular}{c c c cc c cc c c c c c}
        \toprule
        Methods & Metrics & Power & Low{\_}veg & Imp{\_}surf & Car & Fence{\_}hedge & Roof & Fac & Shrub & Tree & $OA$ & $AvgF_1$\\
        \midrule
        \multirow{3}{*}{PointNet \citep{qi2017pointnet}}     & $Pr$  &  0.0 & 67.5 & 90.2 &  0.0 &  0.0 & 73.0 & 10.7 & 28.1 & 48.1 & \multirow{3}{*}{71.2} & \multirow{3}{*}{69.3}\\ 
                                                            & $Re$  &  0.0 & 86.7 & 94.9 &  0.0 &  0.0 & 82.2 &  0.1 & 28.9 & 26.0 &\\                                              
                                                            & $F_1$ &  0.0 & 77.1 & 92.6 &  0.0 &  0.0 & 77.6 &  5.4 & 28.5 & 37.0 &\\                              
        \midrule
        \multirow{3}{*}{PointNet++ \citep{qi2017pointnet++}} & $Pr$  & 76.9 & 84.5 & 99.2 & 60.2 & 24.3 & 92.8 & 43.8 & 26.5 & 62.0 & \multirow{3}{*}{79.1} & \multirow{3}{*}{60.5}\\
                                                            & $Re$  & 40.0 & 78.4 & 97.1 & 50.4 & 14.1 & 81.0 & 38.3 & 39.9 & 80.3\\
                                                            & $F_1$ & 58.5 & 81.4 & 98.1 & 55.3 & 19.2 & 86.9 & 41.0 & 33.2 & 71.2\\
        \midrule
        \multirow{3}{*}{HDA-PointNet++ \citep{huang2020deep}}& $Pr$  & 70.0 & 85.2 & 98.9 & 72.7 & 31.9 & 92.8 & 51.4 & 29.9 & 63.1 & \multirow{3}{*}{81.2} &  \multirow{3}{*}{63.1}\\
                                                            & $Re$  & 58.3 & 85.0 & 99.4 & 65.2 & 6.6  & 83.5 & 21.5 & 45.5 & 75.4\\
                                                            & $F_1$ & 64.2 & \textbf{85.1} & \textbf{99.2} & 68.9 & 19.2 & 88.2 & 36.5 & 37.7 & 69.2\\
        \midrule
        \multirow{3}{*}{PointSIFT \citep{jiang2018pointsift}}& $Pr$  & 100.0 & 83.2 & 88.1 & 86.7 & 64.3 & 93.6 & 64.3 & 38.7 & 76.8 & \multirow{3}{*}{82.7} & \multirow{3}{*}{67.9}\\
                                                            & $Re$  & 1.2 & 77.4 & 94.4 & 63.2 & 12.5 & 92.1 & 54.2 & 45.0 & 86.5 &\\
                                                            & $F_1$ & 50.6 & 80.3 & 91.3 & 74.9 & 38.4 & 92.9 & 59.2 & 41.8 & 81.4 &\\                                                            
        \midrule
        \multirow{3}{*}{Ours} & $Pr$ & 80.2 & 87.1 & 89.2 & 92.3 & 81.0 & 94.3 & 69.6 & 42.7 & 79.3 & \multirow{3}{*}{\textbf{84.5}} & \multirow{3}{*}{\textbf{73.5}}\\
                                                            & $Re$  & 52.5 & 78.5 & 94.3 & 69.1 & 21.4 & 94.9 & 54.6 & 57.1 & 85.0 &\\
                                                            & $F_1$ & \textbf{66.3} & 82.8 & 91.8 & \textbf{80.7} & \textbf{51.2} & \textbf{94.6} & \textbf{62.1} & \textbf{49.9} & \textbf{82.1} &\\
        \bottomrule
    \end{tabular}}
\end{table*}

\begin{figure*}[h!]
	\begin{center}
		\includegraphics[width=\textwidth]{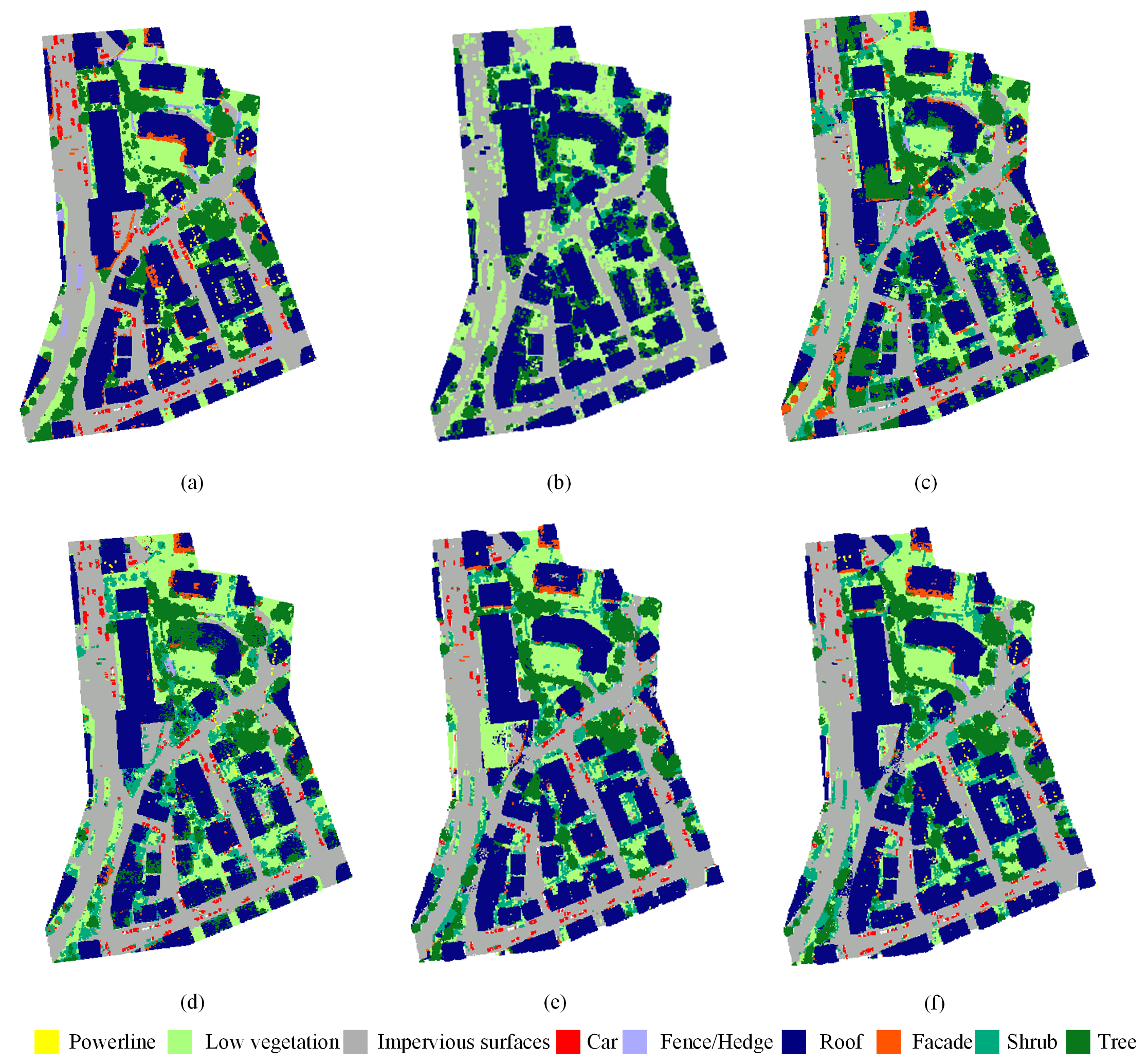}
		\caption{Classification results of test area 1 of the ISPRS benchmark dataset. (a) Ground truth. (b) PointNet. (c) PointNet++. (d) HDA-PointNet++. (e) PointSIFT. (f) Ours.}
		\label{fig:isprs_results1}
    \end{center}
\end{figure*}
\begin{figure*}[ht]
	\begin{center}
		 \includegraphics[width=\textwidth]{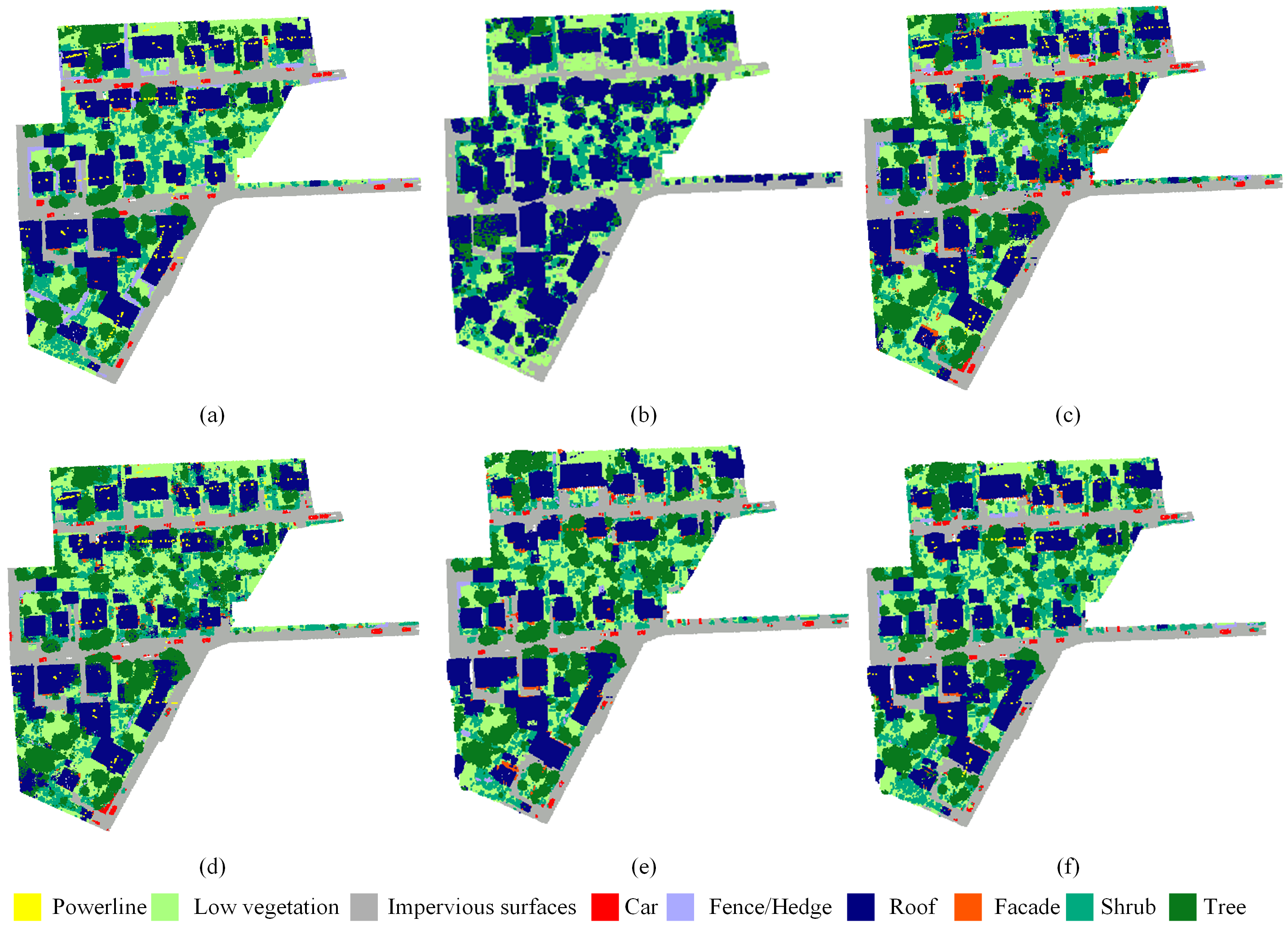}
		\caption{Classification results of test area 2 of the ISPRS benchmark dataset. (a) Ground truth. (b) PointNet. (c) PointNet++. (d) HDA-PointNet++. (e) PointSIFT. (f) Ours.}
		\label{fig:isprs_results2}
    \end{center}
\end{figure*}

\begin{figure*}[htp!]
	\begin{center}
		 \includegraphics[width=\textwidth]{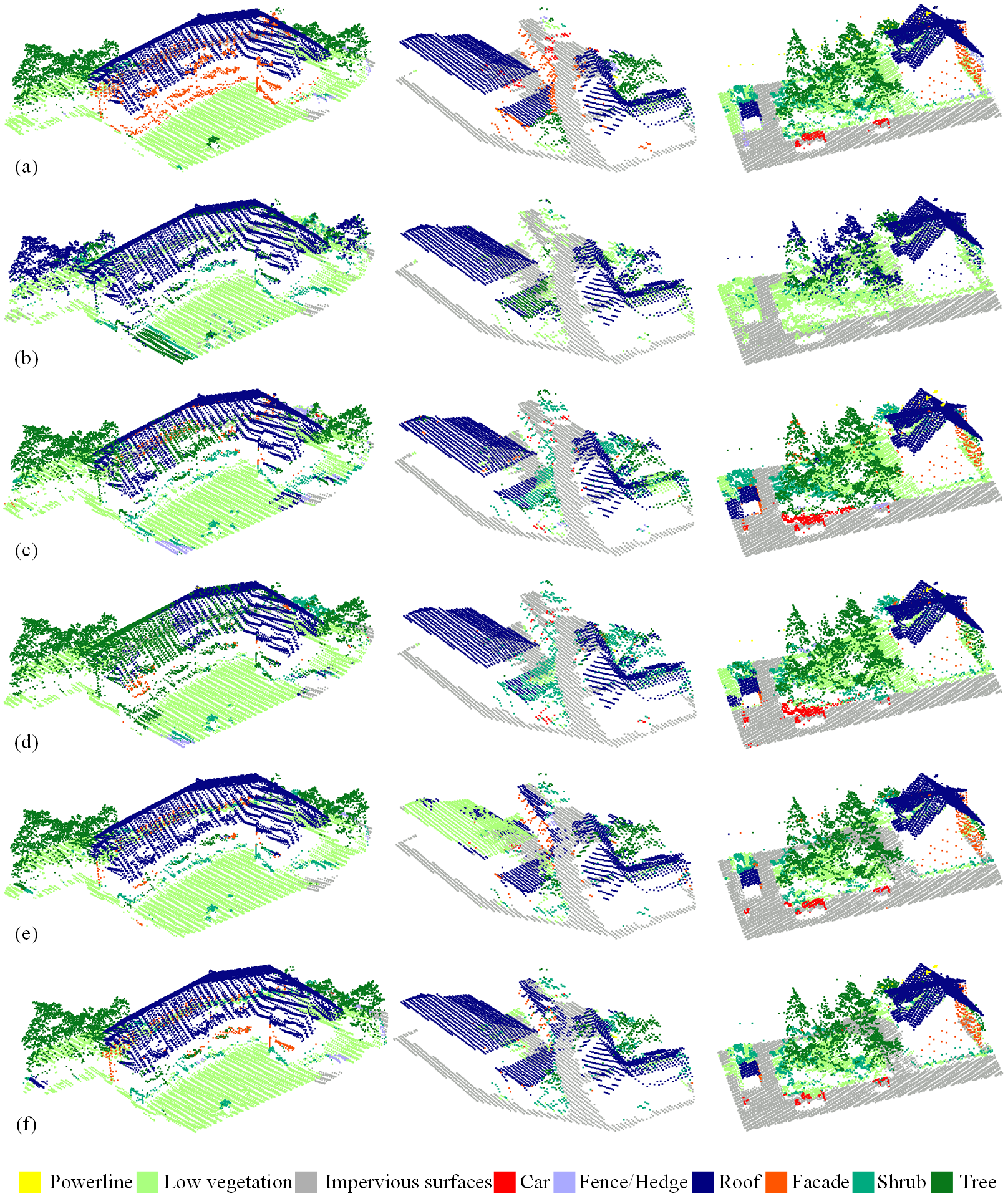}
		\caption{Details of classification results of the ISPRS benchmark dataset. (a) Ground truth. (b) PointNet. (c) PointNet++. (d) HDA-PointNet++. (e) PointSIFT. (f) Ours.}
		\label{fig:isprs_results_d}
    \end{center}
\end{figure*}

In Figs.~\ref{fig:isprs_results1} and ~\ref{fig:isprs_results2}, we provide a visualized classification results. 
As seen from these two figures, it is difficult for the PointNet method to recognize the objects in small scales and with complicated patterns. 
However, when embedding the PointNet as a unit in the multi-scale framework, the PointNet++ method has already provided good performance, indicating the efficacy of the deep neural network when creating productive features. 
It can also be observed that different methods show totally varied output characteristics, although they are all implemented under a similar multi-scale frame from PointNet++. 
We also provide some details for the classification results in Fig.\ref{fig:isprs_results_d}. 
As mentioned before, PointNet++ has already shown good ability in recognizing complicated and fine-grained patterns. 
However, for the low vegetation and trees crowing the building facades, PointNet++ gets a failure when dealing with these cases. 
By contrast, the HDA-PointNet++ method with a hierarchical structure can better handle this issue more robustly and adaptively. 
However, for points of fences, they are all wrongly recognized as low vegetation by HDA-PointNet+ because from the aspect of scales, these two kinds of objects are of quite similar scales. 
As for PointSIFT, it has a strong ability to distinguish details since it constructs local orientation features in different scales. 
As shown by the figure, this method performs better in preserving boundaries between objects. 
However, in a larger scope, it may cause errors. 
For example, it wrong recognized buildings as low vegetation due to the similar local geometric characteristics. 
For our proposed method, although it still shows some mistakes when classifying facades, it produces better results when distinguishing low vegetation, trees, and shrub, compared with other methods. Additionally, our method performs well in classifying small objects, such as powerlines, cars, and fences. 
Simultaneously, our labeled objects reveal sharp and clear boundaries, which provides a potential for further segmentation of individual objects. 
Therefore, it can be concluded that the local attentional spatial encoding and the relation-aware module provide a better-visualized classification map than other methods using a similar PointNet++ frame.

\subsubsection{Error map of the classification results}

\begin{figure*}[ht!]
	\begin{center}
		 \includegraphics[width=0.6\textwidth]{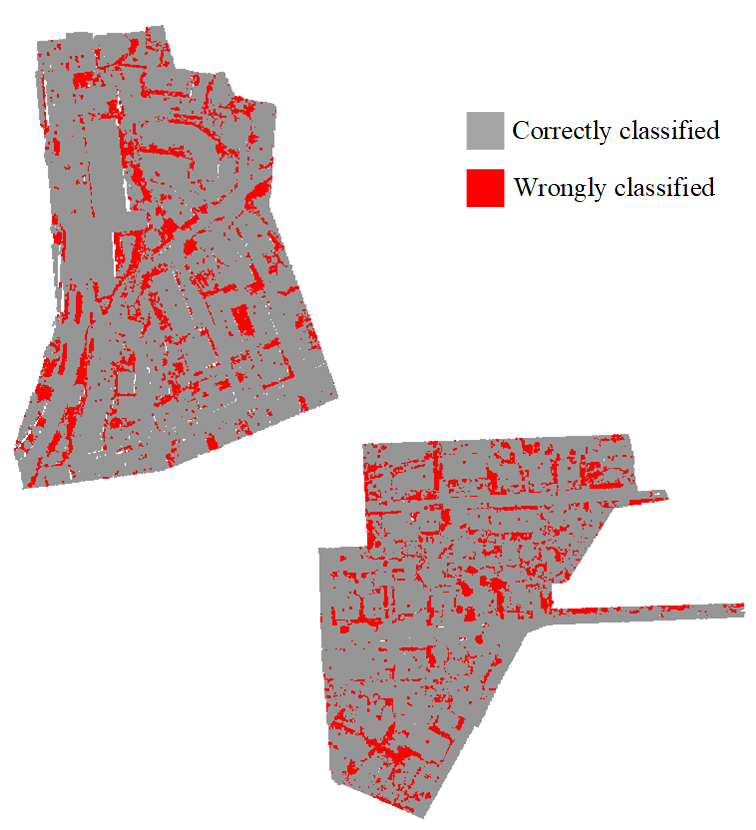}
		\caption{The error map of the classification results of the ISPRS benchmark dataset using our method.}
		\label{fig:isprs_results_errormap}
    \end{center}
\end{figure*}

Fig.~\ref{fig:isprs_results_errormap} illustrates the error map of the classification results using our method on the ISPRS benchmark dataset. 
From the figure, it can be seen that the majority of 3D points can be assigned with correct labels. 
However, there are still some errors in the classification maps. 
With reference to the ground truth provided in Figs.~\ref{fig:isprs_results1} and \ref{fig:isprs_results2}, most of the errors lie in the boundaries between buildings and facades and also the boundaries between vegetation, including low vegetation, tree, and shrub. These parts do indeed share similar geometric characteristics, and the local neighborhood is also similar when the point cloud is sparse. 
Additionally, for the classification of vegetation, the lack of sufficient spectral information may also be one reason that it is hard to distinguish these three categories, especially when the geometric attributes are similar.

\subsubsection{Comparing with results from other published methods}

For further evaluation, we also compare the proposed method with other published methods achieving baseline results, which can be regarded as reference methods, including LUH \citep{niemeyer2016hierarchical}, NANJ2 \citep{zhao2018classifying}, WhuY4 \citep{yang2018segmentation}, RIT{\_}1 \citep{yousefhussien2018multi}, DPE \citep{huang2020deep}, and a geometry attentional network (GANet) \citep{li2020geometry}, DANCE-NET \citep{li2020dance}, and a directionally constrained fully convolutional neural network (D-FCN) \citep{wen2020directionally}. 
The published results of these methods can be checked from the ISPRS 3D semantic labeling dataset website\footnote{http://www2.isprs.org/commissions/comm2/wg4/vaihingen-3d-semantic-labeling.html} or published papers. 
In Table \ref{tab:isprs_baselines}, a comparison of the classification results obtained by these methods is provided and displayed with the aforementioned evaluation metrics. 
LUH utilized a two-layer hierarchical CRF. 
For these two layers, one layer operated directly on points, while the other operated on generated segments using manual features. Apart from LUH, all the other methods are deep learning-based methods. Noted that not all the abovementioned methods directly handle the discrete points. 
NANJ2 and WUY4 methods rely on 2D deep learning strategies. 
The NANJ2 method utilized a 2D CNN to predict labels of ALS point clouds in an urban scenario by learning depth features of multiple scales with several selected attributes, including height, intensity, roughness, and a color vector. 
The WUY4 method had a similar strategy as that uses in the NANJ2 method, wherein the pixel-wise features, generated based on the geometric attribution of 3D points, were fed into a 2D CNN to obtain the classification results.
Unlike the 2D based methods, RIT{\_}1 was implemented directly on 3D points by using a proposed multi-scale 1D fully convolutional architecture. 
The DPE method is a hybrid of conventional manifold learning on the deep features extracted from PointNet++, followed by graph-based optimization.
The GANet proposed a geometry attentional network, which embedded the geometry-aware convolution, elevation attentional modules, and a dense hierarchical architecture in the general framework provided by PointNet++. DANCE-NET proposed a density-aware convolution module for encoding local structural information. D-FCN introduced a directionally constrained fully convolutional neural network, which learned local orientation-aware representation from 2D projected fields. 
\begin{table*}[ht!]
    \centering
    \caption{Comparing with results using the ISPRS benchmark dataset with other published methods  (Values in {\%}). Noted that the highest values in $OA$, $AvgF_1$, and $F_1$ for each category are marked with bold texts. \label{tab:isprs_baselines}}
    \resizebox{\textwidth}{!}
    {\begin{tabular}{c c c cc c cc c c c c c}
        \toprule
        Methods & Metrics & Power & Low{\_}veg & Imp{\_}surf & Car & Fence{\_}hedge & Roof & Fac & Shrub & Tree & $OA$ & $AvgF_1$\\
        \midrule
        \multirow{3}{*}{LUH \citep{niemeyer2016hierarchical}}   & $Pr$  & 67.9 & 83.0 & 91.8 & 86.4 & 49.5 & 97.3 & 52.4 & 34.1 & 87.4 & \multirow{3}{*}{81.6} & \multirow{3}{*}{68.4}\\
                                                                & $Re$  & 53.2 & 72.7 & 90.4 & 63.3 & 25.9 & 91.3 & 60.9 & 73.4 & 79.1 &\\
                                                                & $F_1$ & 59.6 & 77.5 & 91.1 & 73.1 & 34.0 & 94.2 & 56.3 & 46.6 & 83.1 &\\       
        \midrule
        \multirow{3}{*}{NANJ2 \citep{zhao2018classifying}}      & $Pr$  & 62.8 & 90.0 & 89.2 & 83.4 & 50.5 & 95.7 & 47.6 & 45.4 & 88.3 & \multirow{3}{*}{\textbf{85.2}} & \multirow{3}{*}{69.3}\\
                                                                & $Re$  & 61.2 & 87.7 & 93.3 & 55.6 & 34.0 & 91.6 & 38.6 & 72.7 & 77.5 &\\
                                                                & $F_1$ & 62.0 & \textbf{88.8} & 91.2 & 66.7 & 40.7 & 93.6 & 42.6 & \textbf{55.9} & \textbf{82.6} &\\
        \midrule
        
        \multirow{3}{*}{WhuY4 \citep{yang2018segmentation}}               &$Pr$   & 66.5 & 80.6 & 90.4 & 71.0 & 73.0 & 93.1 & 62.4 & 55.2 & 81.9 &\multirow{3}{*}{84.9} &\multirow{3}{*}{69.2}\\
                                                                & $Re$  & 31.2 & 85.0 & 92.4 & 78.9 & 42.5  & 95.6 & 46.2 & 42.4 & 83.7\\
                                                                & $F_1$ & 42.5 & 82.7 & 91.4 & 74.7 & \textbf{53.7} & 94.3 & 53.1 & 47.9 & 82.8\\                                    
        \midrule
        
        \multirow{3}{*}{RIT{\_}1 \citep{yousefhussien2018multi}}   & $Pr$  & 50.4 & 88.0 & 89.6 & 70.1 & 66.5 & 95.2 & 51.4 & 33.4 & 86.0 & \multirow{3}{*}{81.6} & \multirow{3}{*}{63.3}\\
                                                                & $Re$  & 29.8 & 69.8 & 93.6 & 77.0 & 10.4 & 92.9 & 47.4 & 73.4 & 79.3 &\\
                                                                & $F_1$ & 37.5 & 77.9 & 91.5 & 73.4 & 18.0 & 94.0 & 49.3 & 45.9 & 82.5 &\\
        \midrule
        
        \multirow{3}{*}{DPE \citep{huang2020deep}}               &$Pr$   & 77.4 & 88.3 & 99.0 & 83.9 & 34.0 & 92.8 & 64.2 & 31.4 & 66.9 &\multirow{3}{*}{83.2} &\multirow{3}{*}{66.2}\\
                                                                & $Re$  & 58.8 & 84.7 & 99.6 & 66.6 & 4.9  & 89.3 & 24.2 & 47.4 & 78.3\\
                                                                & $F_1$ & 68.1 & 86.5 & \textbf{99.3} & 75.2 & 19.5 & 91.1 & 44.2 & 39.4 & 72.6\\                                                        
        \midrule
        
        \multirow{3}{*}{GANet \citep{li2020geometry}}               &$Pr$   & 73.5 & 85.4 & 89.3 & 81.7 & 58.8 & 93.8 & 74.7 & 43.5 & 83.1 &\multirow{3}{*}{84.5} &\multirow{3}{*}{73.2}\\
                                                                & $Re$  & 77.5 & 78.9 & 94.0 & 74.2 & 35.3  & 95.0 & 52.3 & 57.6 & 82.1\\
                                                                & $F_1$ & \textbf{75.4} & 82.0 & 91.6 & 77.8 & 44.2 & 94.4 & 61.5 & 49.6 & \textbf{82.6}\\                                    
        \midrule
        
        \multirow{3}{*}{DANCE-NET \citep{li2020dance}}               &$Pr$   & 66.8 & 77.1 & 94.6 & 74.5 & 28.3 & 93.7 & 53.5 & 58.0 & 83.4 &\multirow{3}{*}{83.9} &\multirow{3}{*}{71.2}\\
                                                                & $Re$  & 70.0 & 86.6 & 91.0 & 80.2 & 60.6  & 94.2 & 69.0 & 39.8 & 79.5\\
                                                                & $F_1$ & 68.4 & 81.6 & 92.8 & 77.2 & 38.6 & 93.9 & 60.2 & 47.2 & 81.4\\                                    
        \midrule
        
        \multirow{3}{*}{D-FCN \citep{wen2020directionally}}               &$Pr$   & 71.8 & 83.6 & 92.7 & 86.2 & 62.1 & 95.4 & 63.7 & 36.4 & 76.7 &\multirow{3}{*}{82.2} &\multirow{3}{*}{70.7}\\
                                                                & $Re$  & 69.0 & 77.1 & 90.2 & 71.4 & 26.4 & 90.7 & 57.5 & 62.6 & 82.2\\
                                                                & $F_1$ & 70.4 & 80.2 & 91.4 & 78.1 & 37.0 & 93.0 & 60.5 & 46.0 & 79.4\\                                    
        \midrule
        
       \multirow{3}{*}{Ours} & $Pr$ & 80.2 & 87.1 & 89.2 & 92.3 & 81.0 & 94.3 & 69.6 & 42.7 & 79.3 & \multirow{3}{*}{84.5} & \multirow{3}{*}{\textbf{73.5}}\\
                                                            & $Re$  & 52.5 & 78.5 & 94.3 & 69.1 & 21.4 & 94.9 & 54.6 & 57.1 & 85.0 &\\
                                                            & $F_1$ & 66.3 & 82.8 & 91.8 & \textbf{80.7} & 51.2 & \textbf{94.6} & \textbf{62.1} & 49.9 & 82.1 &\\
        \bottomrule
    \end{tabular}}
\end{table*}
When compared with other aforementioned methods, one major strength of our method is that our method ranks first among all the listed state-of-the-art methods in terms of $AvgF_1$ of 73.5{\%}. Its $OA$ is the same as the value of GANet, but its $AvgF_1$ is 0.3{\%} higher than GANet. 
Besides, compared to NANJ2 and WhuY4, although we did not achieve higher $OA$, the $AvgF_1$ of our method is higher by 4.2{\%}, and 4.3{\%}, respectively. 
Additionally, our method is an end-to-end method that directly works on 3D points compared with these two methods. 
Compared with the other 3D methods, our method has a higher $OA$ and $AvgF_1$.

\subsection{Classification results of LASDU dataset}

\subsubsection{Implementation details}

Considering that the LASDU dataset has a similar point density to the ISPRS benchmark dataset, we conducted similar division procedures on the LASDU dataset to generate the data for the training and validation process. 
The whole training data was split into several 100$\times$100 large blocks without overlaps. Then, 90{\%} of blocks were used for training, and the other 10{\%} blocks were treated as validation data. 
All these blocks were subdivided into 25 $\times$ 25 subblocks with 12.5-m overlaps in both $x-$ and $y-$directions. 
Finally, before putting all these small blocks into the network for the training process, the points were resampled from each point set to meet the input requirement. 
The number of points as the input size was set as 4096.  
Additionally, the hyperparameters set for the training process was also similar. 
The batch size, the initial learning rate, the decay rate, and the max epoch were set to 4, 0.001, 0.7, and 1000, respectively. 
The network was trained using the same GPU and virtual environment.

\subsubsection{Results using LASDU dataset}

The same as the comparison in the evaluation using the ISPRS benchmark dataset, we also compare the results with the other method under the framework of PointNet++. 
Table \ref{tab:lasdu_results} lists the classification results of four different PointNet++ based methods, consisting of PointNet, PointNet++, HDA-PointNet++, and PointSIFT. 
Since the urban scene presented in the LASDU dataset is less complicated than the ISPRS benchmark dataset, the performance of PointNet is slightly better. 
At least, most urban objects can be distinguished. 
Compared to the baseline method, PointNet++, using the proposed method, $OA$ and $AvgF_1$ are incremented by 3.5{\%} and 4.7{\%}. 
By stacking features from multiple scales, recognizing small objects and distinguishing objects from different scales can be strengthened using HDA-PointNet++. 
However, compared to our method, $OA$ and $AvgF_1$ are still lower by 1.9{\%} and 2.6{\%}. 
Additionally, compared to PointSIFT, which considers the orientation and scale information, our method still performs better by considering additional elevation information and long-term dependencies. Additionally, when it comes to the classification accuracies, it can be seen from the table that our method achieves the best results in most categories among the aforementioned methods. 

\begin{figure*}[hp]
    \centering
    \includegraphics[width=1\textwidth]{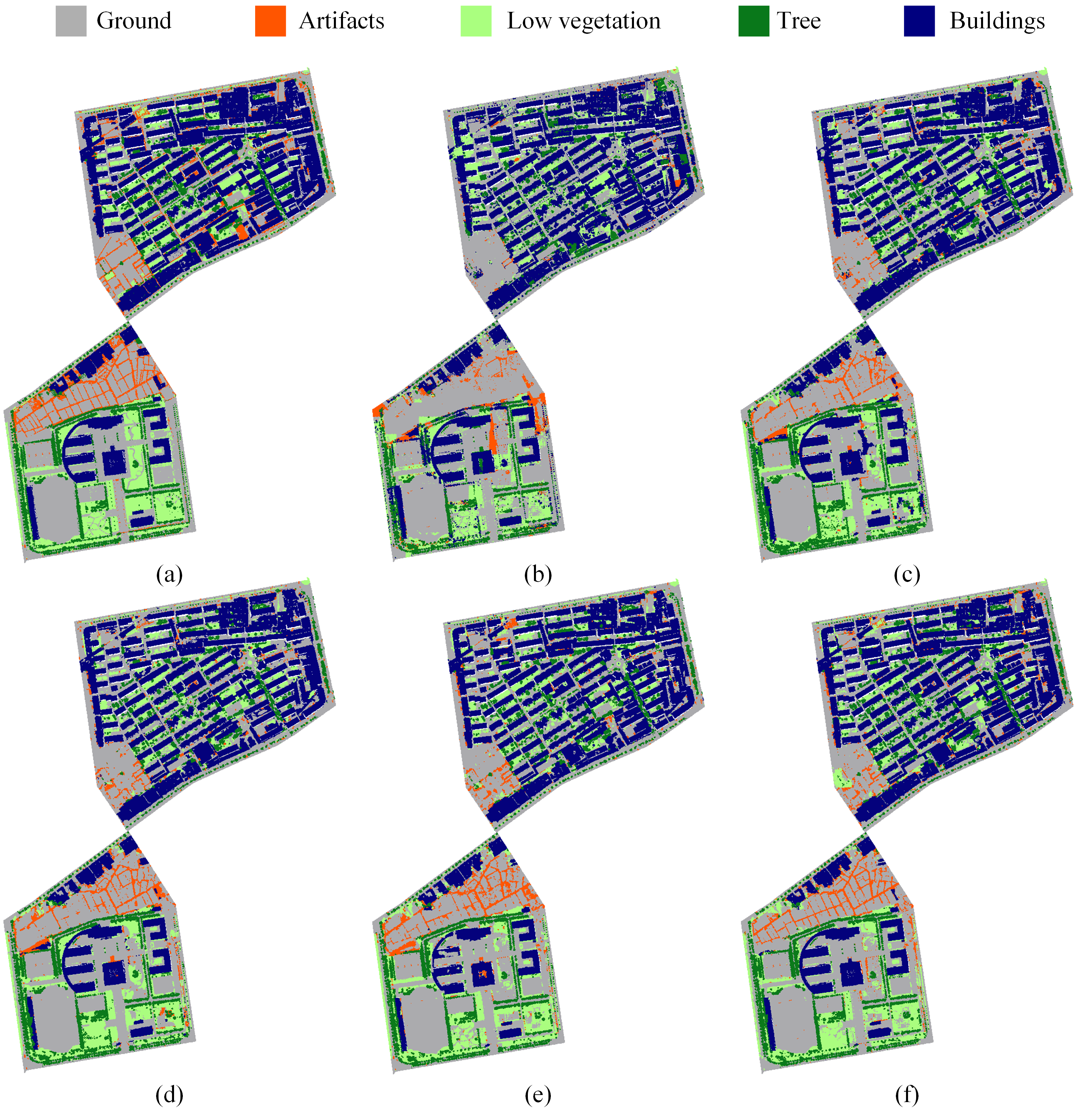}
    \caption{Classification results using the LASDU dataset with different semantic labeling methods. (a) Ground truth, (b) PointNet. (c) PointNet++. (d) HDA-PointNet++. (e) PointSIFT. (f) Ours.}
    \label{fig:Lasdu_results}
\end{figure*}

\begin{figure*}[htp!]
	\begin{center}
		 \includegraphics[width=0.9\textwidth]{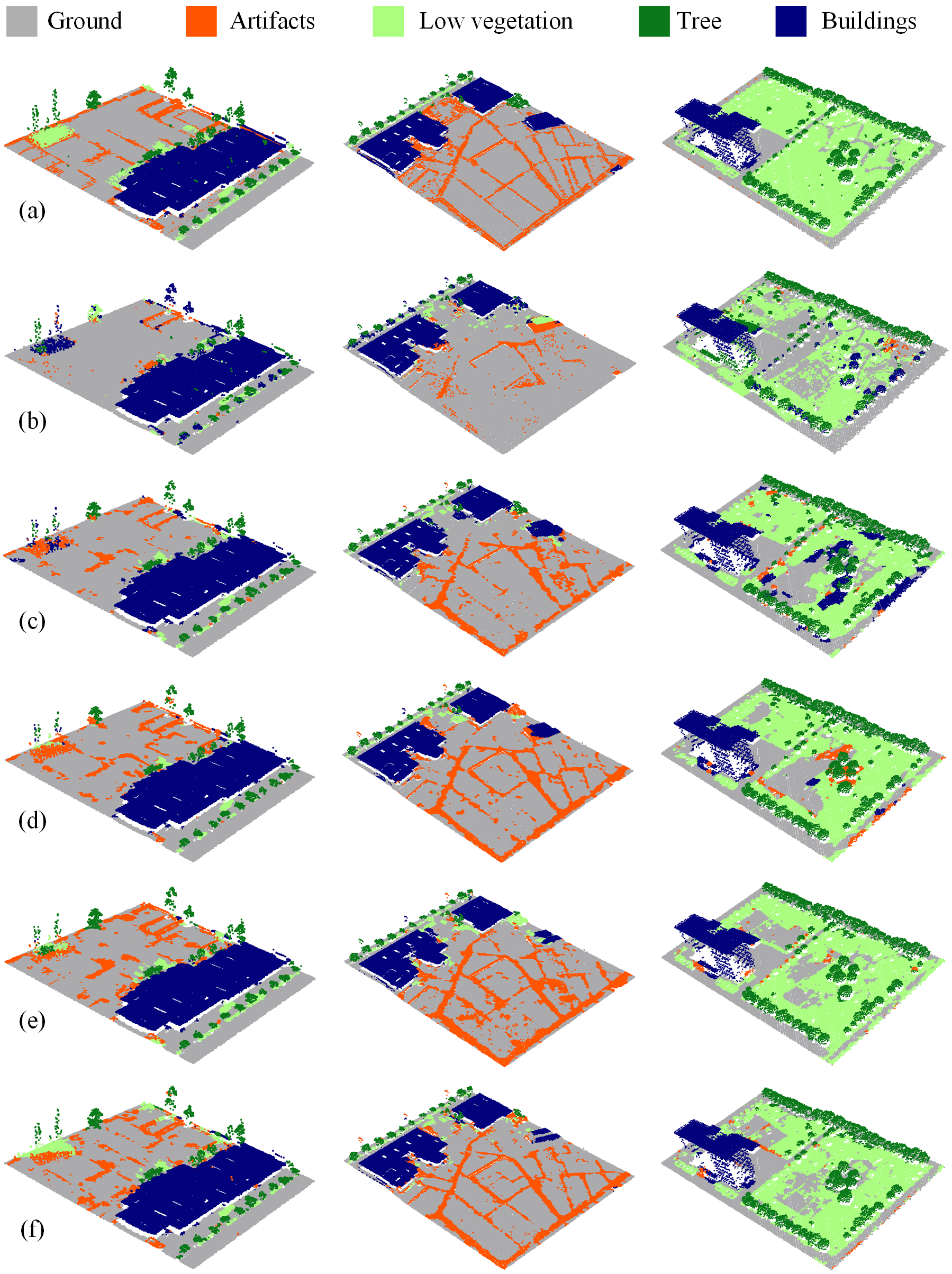}
		\caption{Details of classification results using the LASDU dataset. (a) Ground truth. (b) PointNet. (c) PointNet++. (d) HDA-PointNet++. (e) PointSIFT. (f) Ours.}
		\label{fig:Lasdu_results_d}
    \end{center}
\end{figure*}

\begin{table*}[ht!]
    \centering
    \caption{Comparison of classification results using the LASDU dataset with different PointNet++ methods (all values are in {\%}). Noted that the highest values in $OA$, $AvgF_1$, and $F_1$ for each category are marked with bold texts.\label{tab:lasdu_results}}
    \resizebox{\textwidth}{!}{
    \begin{tabular}{c c c c c c c c c}
        \toprule
        Methods                     & Metrics & Artifacts & Buildings & Ground & Low{\_}veg & Trees & $OA$                  & $\text{Avg}F_1$\\
        \midrule
        \multirow{3}{*}{PointNet \citep{qi2017pointnet}} & $Pre$   & 16.8   & 87.4   & 80.6   & 56.9   & 65.5    & \multirow{3}{*}{77.5} &\multirow{3}{*}{59.3}\\
                                    & $Rec$   & 9.7   & 85.1   & 92.1   & 45.0   & 54.4    & & \\
                                    & $F1$    & 13.2   & 86.2   & 86.3   & 51.0   & 59.9    & & \\
        \midrule
        \multirow{3}{*}{PointNet++ \citep{qi2017pointnet++}} & $Pre$   & 34.7   & 91.4   & 85.8   & 66.8   & 78.4    & \multirow{3}{*}{82.8} &\multirow{3}{*}{71.0}\\
                                    & $Rec$   & 27.8   & 89.8   & 89.6   & 59.6   & 85.6    & & \\
                                    & $F1$    & 31.3   & 90.6   & 87.7   & 63.2   & 82.0    & & \\
        \midrule
        \multirow{3}{*}{HDA-PointNet++ \citep{huang2020deep}}        & $Pre$   & 38.7   & 95.9   & 85.7   & 69.5   & 80.0    & \multirow{3}{*}{84.4} &\multirow{3}{*}{73.2}\\
                                    & $Rec$   & 35.1   & 90.4   & 91.8   & 61.0   & 84.5    & & \\
                                    & $F1$    & 36.9   & 93.2   & 88.7   & \textbf{65.2}   & 82.2    & & \\
        \midrule
        \multirow{3}{*}{PointSIFT \citep{jiang2018pointsift}} & $Pre$   & 39.6   & 96.8   & 85.7   & 68.6   & 84.8    & \multirow{3}{*}{84.9} &\multirow{3}{*}{74.2}\\
                                    & $Rec$   & 36.5   & 91.8   & 91.8   & 60.2   & 86.2    & & \\
                                    & $F1$    & 38.0   & 94.3   & 88.8   & 64.4   & 85.5    & & \\
        \midrule
        \multirow{3}{*}{Ours} & $Pre$   & 45.4   & 97.6   & 87.1   & 70.4   & 83.2    & \multirow{3}{*}{\textbf{86.3}} &\multirow{3}{*}{\textbf{75.8}}\\
                                    & $Rec$   & 40.0   & 94.1   & 93.3   & 58.9   & 88.4    & & \\
                                    & $F1$    & \textbf{42.7}   & \textbf{95.8}   & \textbf{90.2}   & 64.6   & \textbf{85.6}    & & \\
        \bottomrule
    \end{tabular}}
\end{table*}
As shown in Figs.~\ref{fig:Lasdu_results} and~\ref{fig:Lasdu_results_d}, using the proposed method, most areas of our interest, such as buildings, artifacts, and vegetation, can be correctly classified. 
Especially for artifacts, it is clear that our method performs the best among all the aforementioned methods in terms of recognizing artifacts and preserving their boundaries between ground. 
However, our proposed method also meet some problems in classifying low vegetation and ground for some specific areas.

\section{Discussion}\label{Sec:DIS}

\subsection{Ablation study}

In order to investigate the effectiveness of each individual module in the network, we conduct the following ablation studies. 
All the experiments were conducted on the ISPRS benchmark dataset. 
The detailed design of the ablation studies and the evaluation results are provided in the following. 
\subsubsection{Effectiveness of LoSDA module}
To vindicate the LoSDA module and explore the effectiveness of each module in LoSDA, we trained five models using the same network architecture with different local spatial encoding modules, namely only SDE (model \textbf{A}), only DFE (model \textbf{B}), the combination of SDE and DFE (model \textbf{C}), the combination of SDE, DFE and EDE (model \textbf{D}), the full local spatial encoding module using the final combination of SDE, DFE, and EDE with attention pooling (model \textbf{E}). 
All the experimental results are listed in Table \ref{tab:results_isprs_losda}.
\begin{table*}[ht!]
    \centering
    \caption{Comparing with results using different local spatial encoding methods using the ISPRS benchmark dataset (Values in {\%}). Noted that the highest values in $OA$, $AvgF_1$, and $F_1$ for each category are marked with bold texts. \label{tab:results_isprs_losda}}
    \resizebox{\textwidth}{!}
    {\begin{tabular}{c c c cc c cc c c c c c}
        \toprule
        Methods & Metrics & Power & Low{\_}veg & Imp{\_}surf & Car & Fence{\_}hedge & Roof & Fac & Shrub & Tree & $OA$ & $AvgF_1$\\
        \midrule
        \multirow{3}{*}{\textbf{A}} & $Pr$  & 76.7 & 82.7 & 89.7 & 86.0 & 57.7 & 95.0 & 67.2 & 40.1 & 71.5 & \multirow{3}{*}{81.8} & \multirow{3}{*}{67.6}\\
                                                            & $Re$  & 23.0 & 80.3 & 91.3 & 67.5 & 19.2 & 88.2 & 41.7 & 55.5 & 84.4 &\\
                                                            & $F_1$ & 49.8 & 81.5 & 90.5 & 76.8 & 38.4 & 91.6 & 54.5 & \textbf{47.8} & 78.0 &\\
        \midrule
        \multirow{3}{*}{\textbf{B}} & $Pr$  & 100.0 & 83.2 & 88.1 & 86.7 & 64.3 & 93.6 & 64.3 & 38.7 & 76.8 & \multirow{3}{*}{82.7} & \multirow{3}{*}{67.9}\\
                                                            & $Re$  & 1.2 & 77.4 & 94.4 & 63.2 & 12.5 & 92.1 & 54.2 & 45.0 & 86.5 &\\
                                                            & $F_1$ & 50.6 & 80.3 & 91.3 & 74.9 & 38.4 & 92.9 & 59.2 & 41.8 & \textbf{81.4} &\\
        \midrule
        \multirow{3}{*}{\textbf{C}} & $Pr$  & 57.8 & 84.5 & 89.0 & 93.9 & 69.6 & 93.6 & 71.7 & 41.3 & 77.1 & \multirow{3}{*}{83.6} & \multirow{3}{*}{68.9}\\
                                                            & $Re$  & 41.2 & 79.9 & 94.4 & 53.2 & 16.7 & 93.9 & 49.5 & 50.5 & 83.0 &\\
                                                            & $F_1$ & 49.5 & 82.2 & 91.7 & 73.6 & 43.2 & 93.8 & 60.6 & 45.9 & 80.0 &\\
        \midrule
        \multirow{3}{*}{\textbf{D}} & $Pr$  & 80.7 & 83.8 & 89.3 & 96.4 & 69.9 & 93.4 & 74.5 & 43.2 & 77.3 & \multirow{3}{*}{83.9} & \multirow{3}{*}{71.0}\\
                                                            & $Re$  & 41.0 & 81.0 & 93.1 & 61.3 & 13.8 & 94.3 & 50.8 & 48.9 & 85.5 &\\
                                                            & $F_1$ & \textbf{60.8} & \textbf{82.4} & 91.2 & 78.8 & 41.9 & 93.8 & 62.7 & 46.0 & \textbf{81.4} &\\
        \midrule
        \multirow{3}{*}{\textbf{E}} & $Pr$  & 80.6 & 85.4 & 88.8 & 96.6 & 76.4 & 94.6 & 75.2 & 43.0 & 76.4 & \multirow{3}{*}{\textbf{84.1}} & \multirow{3}{*}{\textbf{71.9}}\\
                                                            & $Re$  & 39.5 & 78.7 & 95.4 & 65.9 & 15.9 & 93.7 & 51.0 & 52.4 & 85.9 &\\
                                                            & $F_1$ & 60.1 & 82.1 & \textbf{92.1} & \textbf{80.2} & \textbf{46.2} & \textbf{94.1} & \textbf{63.1} & 47.7 & 81.1 &\\
        \bottomrule
    \end{tabular}}
\end{table*}

\begin{figure*}[ht!]
    \centering
    \includegraphics[width=1\textwidth]{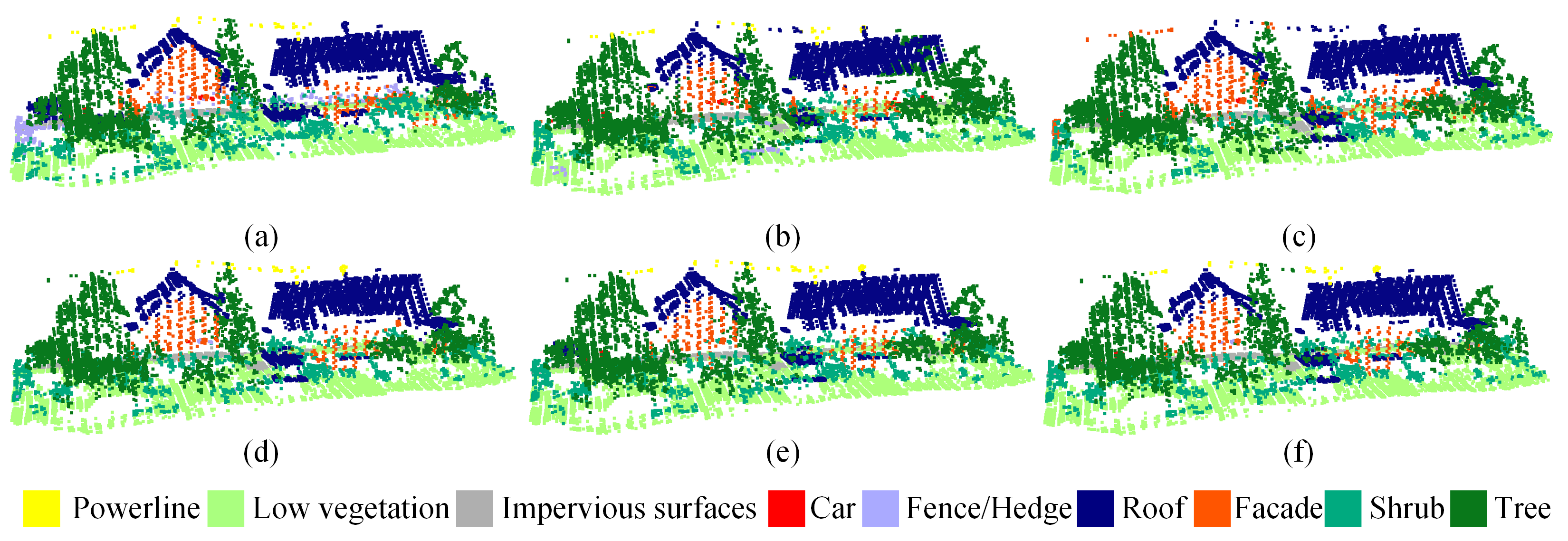}
    \caption{Classification results using different local spatial encoding methods. (a) Ground truth, (b) Model A. (c) Model B. (d) Model C. (e) Model D. (f) Model E.}
    \label{fig:isprs_losda}
\end{figure*}

Results in Table \ref{tab:results_isprs_losda} show that directional features can better present the local geometry compared with features based on local distribution. 
However, since the SDE and the DFE module actually describe the local geometry from different perspectives, we further conduct experiments to test the effectiveness of combining these two local feature embedding modules. 
We can see from the results that the classification results are further improved compared with the results using SDE and DFE individually. 
Additionally, for the ALS point clouds, elevations of points play an essential role. 
It is shown in the table that model D has a higher $OA$ by 0.3{\%} and a higher $AvgF_1$ by 2.1{\%}, compared with model C. 
The results prove that the elevation-based module benefits the description of local geometry. 
The dependencies between local points are further investigated by adding the attention pooling module to the current local spatial encoding module. 
As shown by the results, the classification accuracy is further improved. 
Compared with different combinations of each module in the LoSDA module, we can see that each individual module is useful in learning local feature descriptions. 
Moreover, compared with the baseline method PointSIFT, our local spatial encoding module can better encode the geometric structures in the local scope and is efficient for both large-scale and small-scale categories.
Fig.~\ref{fig:isprs_losda} illustrates the classification results of one selected area in the ISPRS benchmark dataset. 
From the image, it can be seen that the urban objects can be well recognized using the baseline method PointSIFT. 
However, by adding the local spatial encoding, elevation information, and the local dependencies, some small objects can be better distinguished, such as powerline and shrub. 
Moreover, trees, shrubs, and low vegetation, which can be easily misclassified, can also be well recognized. 

\subsubsection{Effectiveness of the GRA module and different modes}

In order to validate the effectiveness of the GRA module and compare the different combination modes of SRA and CRA, we conducted further experiments on the ISPRS benchmark dataset. Namely, based on the LoSDA module and the baseline architecture, we add different modes of the GRA module to the network and evaluate the results. 
It should be mentioned that here the baseline method refers to the network which uses the LoSAD module for local spatial encoding.

\begin{figure*}[ht!]
    \centering
    \includegraphics[width=1\textwidth]{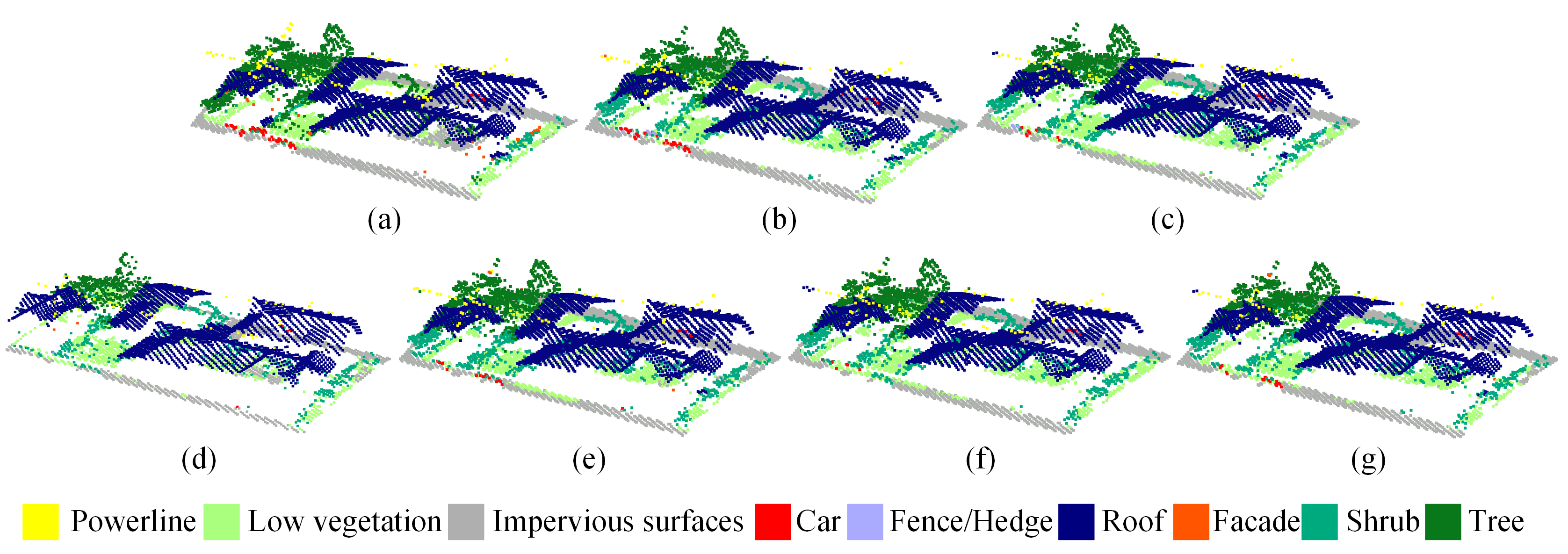}
    \caption{Classification results using different conjurations of the GRA module. (a) Ground truth, (b) Non GRA. (c) Only SRA. (d) Only CRA. (e) Mode 1. (f) Model 2. (g) Mode 3.}
    \label{fig:isprs_gra}
\end{figure*}

\begin{table*}[ht!]
    \centering
    \caption{Comparing with results using different configuration of the GRA module using the ISPRS benchmark dataset (Values in {\%}). Noted that the highest values in $OA$, $AvgF_1$, and $F_1$ for each category are marked with bold texts.\label{tab:results_isprs_gra}}
    \resizebox{\textwidth}{!}
    {\begin{tabular}{c c c cc c cc c c c c c}
        \toprule
        Methods & Metrics & Power & Low{\_}veg & Imp{\_}surf & Car & Fence{\_}hedge & Roof & Fac & Shrub & Tree & $OA$ & $AvgF_1$\\
        \midrule
        \multirow{3}{*}{\textbf{Non GRA}} & $Pr$  & 80.6 & 85.4 & 88.8 & 96.6 & 76.4 & 94.6 & 75.2 & 43.0 & 76.4 & \multirow{3}{*}{84.1} & \multirow{3}{*}{71.9}\\
                                                            & $Re$  & 39.5 & 78.7 & 95.4 & 65.9 & 15.9 & 93.7 & 51.0 & 52.4 & 85.9 &\\
                                                            & $F_1$ & 60.1 & 82.1 & 92.1 & 80.2 & 46.2 & 94.1 & 63.1 & 47.7 & 81.1 &\\
        \midrule
        \multirow{3}{*}{\textbf{Only CRA}} & $Pr$  & 59.9 & 85.6 & 88.4 & 88.3 & 78.6 & 95.2 & 71.1 & 43.2 & 77.4 & \multirow{3}{*}{84.1} & \multirow{3}{*}{72.7}\\
                                                            & $Re$  & 70.0 & 79.1 & 94.7 & 70.9 & 19.5 & 93.4 & 54.2 & 53.5 & 84.6 &\\
                                                            & $F_1$ & 65.2 & 82.4 & 91.6 & 79.6 & 49.1 & 94.3 & 62.6 & 48.3 & 81.0 &\\
        \midrule
        \multirow{3}{*}{\textbf{Only SRA}} & $Pr$  & 76.3 & 87.1 & 88.7 & 94.7 & 74.6 & 94.2 & 71.7 & 41.7 & 81.9 & \multirow{3}{*}{84.3} & \multirow{3}{*}{72.4}\\
                                                            & $Re$  & 49.8 & 78.5 & 95.6 & 55.8 & 22.9 & 94.1 & 51.3 & 62.4 & 81.3 &\\
                                                            & $F_1$ & 63.0 & 82.8 & 92.2 & 75.3 & 48.7 & 94.1 & 61.5 & 52.0 & 81.6 &\\
                                                            
        \midrule
        \multirow{3}{*}{\textbf{Mode 1}} & $Pr$ & 80.2 & 87.1 & 89.2 & 92.3 & 81.0 & 94.3 & 69.6 & 42.7 & 79.3 & \multirow{3}{*}{\textbf{84.5}} & \multirow{3}{*}{\textbf{73.5}}\\
                                                            & $Re$  & 52.5 & 78.5 & 94.3 & 69.1 & 21.4 & 94.9 & 54.6 & 57.1 & 85.0 &\\
                                                            & $F_1$ & 66.3 & 82.8 & 91.8 & 80.7 & \textbf{51.2} & \textbf{94.6} & 62.1 & 49.9 & \textbf{82.1} &\\
        \midrule
        \multirow{3}{*}{\textbf{Mode 2}} & $Pr$  & 68.1 & 85.4 & 90.3 & 93.1 & 76.1 & 94.3 & 70.8 & 43.0 & 77.3 & \multirow{3}{*}{84.3} & \multirow{3}{*}{73.0}\\
                                                            & $Re$  & 66.2 & 80.4 & 93.4 & 67.5 & 19.5 & 93.8 & 54.7 & 53.9 & 85.2 &\\
                                                            & $F_1$ & 67.1 & \textbf{82.9} & 91.9 & 80.3 & 47.8 & 94.1 & 62.8 & 48.4 & 81.2 &\\
        \midrule
        \multirow{3}{*}{\textbf{Mode 3}} & $Pr$  & 78.9 & 86.8 & 89.5 & 93.5 & 77.0 & 94.8 & 74.6 & 41.3 & 78.6 & \multirow{3}{*}{84.3} & \multirow{3}{*}{73.4}\\
                                                            & $Re$  & 56.0 & 79.1 & 95.1 & 68.0 & 19.5 & 93.8 & 53.0 & 59.1 & 82.3 &\\
                                                            & $F_1$ & \textbf{67.4} & \textbf{82.9} & \textbf{92.3} & \textbf{80.8} & 48.3 & 94.3 & \textbf{63.8} & \textbf{50.2} & 80.4 &\\
        \bottomrule
    \end{tabular}}
\end{table*}
All the results using different configurations of the GRA modules are listed in Table \ref{tab:results_isprs_gra}. 
As can be seen from the table, it is clear that the relation-aware attentional module can provide an improvement compared with the baseline method. 
In detail, by adding the CRA module, the $AvgF_1$ experiences an increase by 0.8{\%}. 
The use of the SRA module produces an increment of $OA$ by 0.2{\%} and $AvgF_1$ by 0.5{\%}. 
Besides, by combining the SRA and CRA modules, the performance of our proposed method is further augmented. 
By using the configuration of Mode 1, the results are improved with an increase of $OA$ by 0.4{\%} and $AvgF_1$ by 1.4{\%}. 
Mode 2 and Mode 3 also provide improvement for the classification results. 
It should be mentioned that Mode 1 outperforms all the other configurations of the GRA module, producing the highest $OA$ with 84.5{\%} and $AvgF_1$ with 73.5{\%}. 
In Fig.~\ref{fig:isprs_gra}, a detailed illustration of the classification results using different configurations of the GraNet is provided. 
By using the effective local spatial encoding module, most of the area has been correctly classified, including the roofs, impervious surfaces, and trees. 
By stacking the GRA module, only some small areas are corrected. 
However, even using the GRA module, it is still hard to distinguish low vegetation, trees, and shrub. 
The area shown under the roof is wrongly classified as impervious surfaces in all the test models.

\subsection{Comparing with results using heights above ground}
Since many pieces of research generate heights above ground (HaG) as input for the network \citep{yousefhussien2018multi} or use features generated from the normalized height \citep{niemeyer2014contextual}, we conducted further experiments using HaG instead of exact elevations to test the effect of using HaG. The results of GraNet using exact elevations and HaG are shown in Table \ref{tab:results_isprs_hag}. From the table, we can see that the $OA$ can be improved by using HaG. However, when considering the classification accuracies of each individual category, only the classification of low vegetation and impervious surfaces is improved. Most categories have lower classification accuracies. In addition, the $AvgF_1$ drops by 2.3{\%}. It shows that using HaG can not really improve the classification performance. Moreover, $OA$ is not a perfect index when comparing the classification performance, and it is easily biased when the numbers of categories in a dataset are strongly unbalanced. It would be better to consider other evaluation indexes for a fair comparison.
\begin{table*}[ht!]
    \centering
    \caption{Comparing with results using exact elevations and HaG using the ISPRS benchmark dataset (Values in {\%}). Noted that the highest values in $OA$, $AvgF_1$, and $F_1$ for each category are marked with bold texts. \label{tab:results_isprs_hag}}
    \resizebox{\textwidth}{!}
    {\begin{tabular}{c c c cc c cc c c c c c}
        \toprule
        Methods & Metrics & Power & Low{\_}veg & Imp{\_}surf & Car & Fence{\_}hedge & Roof & Fac & Shrub & Tree & $OA$ & $AvgF_1$\\
        \midrule
        \multirow{3}{*}{\textbf{GraNet}} & $Pr$ & 80.2 & 87.1 & 89.2 & 92.3 & 81.0 & 94.3 & 69.6 & 42.7 & 79.3 & \multirow{3}{*}{84.5} & \multirow{3}{*}{\textbf{73.5}}\\
                                                            & $Re$  & 52.5 & 78.5 & 94.3 & 69.1 & 21.4 & 94.9 & 54.6 & 57.1 & 85.0 &\\
                                                            & $F_1$ & \textbf{66.3} & 82.8 & 91.8 & \textbf{80.7} & \textbf{51.2} & \textbf{94.6} & \textbf{62.1} & \textbf{49.9} & \textbf{82.1} &\\
        \midrule
        
        \multirow{3}{*}{\textbf{GraNet using HaG}} & $Pr$  & 76.1 & 90.0 & 98.5 & 83.1 & 61.0 & 94.0 & 63.2 & 40.0 & 80.2 & \multirow{3}{*}{\textbf{86.7}} & \multirow{3}{*}{71.2}\\
                                                            & $Re$  & 38.8 & 87.0 & 98.9 & 71.4 & 17.1 & 93.2 & 47.7 & 59.0 & 81.6 &\\
                                                            & $F_1$ & 57.5 & \textbf{88.5} & \textbf{98.7} & 77.2 & 39.1 & 93.6 & 55.4 & 49.5 & 80.86 &\\

        \bottomrule
    \end{tabular}}
\end{table*}
\subsection{Complexity and runtime analysis}

To evaluate the complexity and training efficiency of our model, we list the numbers of parameters (params) and running time of eight networks in Tabel \ref{tab:results_complexity}, including two baseline methods, the combination of PointNet++ structure and the LoSDA module, and five different modes of GraNet. 
The input size for all the network is set to 4096. 
It should be mentioned that the running time refers to the time used for one epoch during the training process, including the time used for both training and validation. 
It is clear that the number of parameters is not increased, and the training process is faster using the LoSDA module compared with PointSIFT. 
However, after stacking the GRA modules, the complexity of the model increases dramatically, especially when stacking the CRA module. 
For the different configurations of the GRA module, no matter the parallel or serial configurations, the number of parameters and the running time does not vary too much. 
In general, the complexity of GraNet is higher than the baseline methods. 
The increase of the classification accuracy may partially arise from the increase of the network complexity. 
However, since the training dataset is not a dataset with a large data amount, apart from the PointNet++, the running time of all the other networks varies in a small range. 
\begin{table*}[ht!]
    \centering
    \caption{The number of parameters and running time of different network models. \label{tab:results_complexity}}
    \resizebox{0.6\textwidth}{!}
    {\begin{tabular}{c c c }
        \toprule
        Method & Params (Millions) & Running time (s)\\
        \midrule
        PointNet++ & 0.97 & 32\\
        \midrule
        PointSIFT & 13.56 & 65\\
        \midrule
        PointNet++ {\&} LoSDA & 9.58 & 52\\
        \midrule
        GraNet (only CRA) & 64.82 & 64\\
        \midrule
        GraNet (only SRA) & 11.56 & 64\\
        \midrule
        GraNet mode 1 & 66.81 & 78\\
        \midrule
        GraNet mode 2 & 66.81 & 76\\
        \midrule
        GraNet mode 3 & 66.98 & 78\\
        \bottomrule
    \end{tabular}}
\end{table*}

\section{Conclusions}\label{Sec:CO}

In this work, we propose a novel neural network, GraNet, focusing on ALS point cloud classification, which investigates the way of constructing efficient local neighborhood representation and the importance of long-term dependencies provided by relation-aware modules. 
In GraNet, the local geometric description and local dependencies are first learned using a local spatial discrepancy attention convolution module. 
In LoSDA, the orientation information, spatial distribution, and elevation differences are fully considered by stacking several local spatial geometric learning modules, and the local dependencies are embedded by using an attention pooling module. 
Second, the investigation of the importance of global relation-aware attention is achieved using two modules: the spatial relation-aware attention module and the channel relation-aware attention module. 
Additionally, we also investigate the different configurations of these two relation-aware modules. 
Third, the local spatial encoding module and the relation-aware module are embedded in the multi-scale network architecture to further consider the scale changes in large urban areas.

Experiments were carried out using the ALS dataset to evaluate the performance of our proposed neural network. 
Experimental results demonstrate that our method achieves an $OA$ of 84.5{\%} and $AvgF_1$ of 73.5{\%} on the ISPRS benchmark dataset for classifying points of nine classes of objects and outperforms other commonly used advanced point-based strategies. 
Besides, comparative results also validate the feasibility of using the orientation information, the elevation information, and the local dependencies achieved by local attention pooling to enhance the feature importance, and the importance of considering long-term relations in complicated scenes, especially for urban scenes. 
In the future, we will adopt the proposed multi-scale frame and relation-based strategy to a pre-clustered point structure, which can considerably decrease the processing time and simultaneously retain the superiority of the multi-scale point-based structure when preserving accurate boundaries of objects.

\section*{Acknowledgment}

This research is supported by the China Scholarship Council. This work was carried out within the frame of Leonhard Obermeyer Center (LOC) at Technische Universit{\"a}t M{\"u}nchen (TUM) [www.loc.tum.de]. The authors would like to thank Institute of Tibetan Plateau Research, Chinese Academy of Sciences, China, as well as the HiWATER project, to provide the original ALS point clouds.




%


\bibliographystyle{elsarticle-harv}
\bibliography{mybibfile}

\begin{thebibliography}{79}
\expandafter\ifx\csname natexlab\endcsname\relax\def\natexlab#1{#1}\fi
\expandafter\ifx\csname url\endcsname\relax
  \def\url#1{\texttt{#1}}\fi
\expandafter\ifx\csname urlprefix\endcsname\relax\def\urlprefix{URL }\fi

\bibitem[{Alba et~al.(2006)Alba, Fregonese, Prandi, Scaioni, and
  Valgoi}]{alba2006structural}
Alba, M., Fregonese, L., Prandi, F., Scaioni, M., Valgoi, P., 2006. Structural
  monitoring of a large dam by terrestrial laser scanning. International
  Archives of Photogrammetry, Remote Sensing and Spatial Information Sciences
  36~(5), 6.

\bibitem[{Armeni et~al.(2016)Armeni, Sener, Zamir, Jiang, Brilakis, Fischer,
  and Savarese}]{armeni20163d}
Armeni, I., Sener, O., Zamir, A.~R., Jiang, H., Brilakis, I., Fischer, M.,
  Savarese, S., 2016. 3d semantic parsing of large-scale indoor spaces. In:
  Proceedings of the IEEE Conference on Computer Vision and Pattern
  Recognition. pp. 1534--1543.

\bibitem[{Bosch{\'{e}} et~al.(2015)Bosch{\'{e}}, Ahmed, Turkan, Haas, and
  Haas}]{bosche2015value}
Bosch{\'{e}}, F., Ahmed, M., Turkan, Y., Haas, C.~T., Haas, R., 2015. The value
  of integrating scan-to-bim and scan-vs-bim techniques for construction
  monitoring using laser scanning and bim: The case of cylindrical mep
  components. Automation in Construction 49, 201--213.

\bibitem[{Boulch et~al.(2018)Boulch, Guerry, Le~Saux, and
  Audebert}]{boulch2018snapnet}
Boulch, A., Guerry, J., Le~Saux, B., Audebert, N., 2018. Snapnet: 3d point
  cloud semantic labeling with 2d deep segmentation networks. Computers \&
  Graphics 71, 189--198.

\bibitem[{Chan and Paelinckx(2008)}]{chan2008evaluation}
Chan, J. C.~W., Paelinckx, D., 2008. Evaluation of random forest and adaboost
  tree-based ensemble classification and spectral band selection for ecotope
  mapping using airborne hyperspectral imagery. Remote Sensing of Environment
  112~(6), 2999--3011.

\bibitem[{Chehata et~al.(2009)Chehata, Guo, and Mallet}]{chehata2009airborne}
Chehata, N., Guo, L., Mallet, C., 2009. Airborne lidar feature selection for
  urban classification using random forests. International Archives of
  Photogrammetry, Remote Sensing and Spatial Information Sciences 38~(Part 3),
  W8.

\bibitem[{Chen et~al.(2017)Chen, Ma, Wan, Li, and Xia}]{chen2017multi}
Chen, X., Ma, H., Wan, J., Li, B., Xia, T., 2017. Multi-view 3d object
  detection network for autonomous driving. In: Proceedings of the IEEE
  Conference on Computer Vision and Pattern Recognition. pp. 1907--1915.

\bibitem[{Clode and Rottensteiner(2005)}]{clode2005classification}
Clode, S., Rottensteiner, F., 2005. Classification of trees and powerlines from
  medium resolution airborne laserscanner data in urban environments. In:
  Proceedings of the APRS Workshop on Digital Image Computing (WDIC), Brisbane,
  Australia. Vol.~21.

\bibitem[{Cramer(2010)}]{cramer2010dgpf}
Cramer, M., 2010. The dgpf-test on digital airborne camera evaluation--overview
  and test design. Photogrammetrie-Fernerkundung-Geoinformation 2010~(2),
  73--82.

\bibitem[{Dai et~al.(2017)Dai, Chang, Savva, Halber, Funkhouser, and
  Nie{\ss}ner}]{dai2017scannet}
Dai, A., Chang, A.~X., Savva, M., Halber, M., Funkhouser, T., Nie{\ss}ner, M.,
  2017. Scannet: Richly-annotated 3d reconstructions of indoor scenes. In:
  Proceedings of the IEEE Conference on Computer Vision and Pattern
  Recognition. pp. 5828--5839.

\bibitem[{Engelcke et~al.(2017)Engelcke, Rao, Wang, Tong, and
  Posner}]{engelcke2017vote3deep}
Engelcke, M., Rao, D., Wang, D.~Z., Tong, C.~H., Posner, I., 2017. Vote3deep:
  Fast object detection in 3d point clouds using efficient convolutional neural
  networks. In: IEEE International Conference on Robotics and Automation. IEEE,
  pp. 1355--1361.

\bibitem[{Geiger et~al.(2012)Geiger, Lenz, and Urtasun}]{geiger2012we}
Geiger, A., Lenz, P., Urtasun, R., 2012. Are we ready for autonomous driving?
  the kitti vision benchmark suite. In: IEEE Conference on Computer Vision and
  Pattern Recognition. IEEE, pp. 3354--3361.

\bibitem[{{Ghamisi} and {Höfle}(2017)}]{ghamisi2017}
{Ghamisi}, P., {Höfle}, B., 2017. Lidar data classification using extinction
  profiles and a composite kernel support vector machine. IEEE Geoscience and
  Remote Sensing Letters 14~(5), 659--663.

\bibitem[{Gorgens et~al.(2017)Gorgens, Valbuena, and
  Rodriguez}]{gorgens2017method}
Gorgens, E.~B., Valbuena, R., Rodriguez, L. C.~E., 2017. A method for
  optimizing height threshold when computing airborne laser scanning metrics.
  Photogrammetric Engineering \& Remote Sensing 83~(5), 343--350.

\bibitem[{Guo et~al.(2015)Guo, Huang, Zhang, and Sohn}]{guo2015classification}
Guo, B., Huang, X., Zhang, F., Sohn, G., 2015. Classification of airborne laser
  scanning data using jointboost. ISPRS Journal of Photogrammetry and Remote
  Sensing 100, 71--83.

\bibitem[{Hackel et~al.(2017)Hackel, Savinov, Ladicky, Wegner, Schindler, and
  Pollefeys}]{hackel2017isprs}
Hackel, T., Savinov, N., Ladicky, L., Wegner, J.~D., Schindler, K., Pollefeys,
  M., 2017. {SEMANTIC3D.NET: A new large-scale point cloud classification
  benchmark}. In: ISPRS Annals of the Photogrammetry, Remote Sensing and
  Spatial Information Sciences. Vol. IV-1-W1. pp. 91--98.

\bibitem[{Hebel et~al.(2013)Hebel, Arens, and Stilla}]{hebel2013change}
Hebel, M., Arens, M., Stilla, U., 2013. Change detection in urban areas by
  object-based analysis and on-the-fly comparison of multi-view als data. ISPRS
  Journal of Photogrammetry and Remote Sensing 86, 52--64.

\bibitem[{Hu et~al.(2018)Hu, Gu, Zhang, Dai, and Wei}]{hu2018relation}
Hu, H., Gu, J., Zhang, Z., Dai, J., Wei, Y., 2018. Relation networks for object
  detection. In: Proceedings of the IEEE Conference on Computer Vision and
  Pattern Recognition. pp. 3588--3597.

\bibitem[{Hu et~al.(2020)Hu, Yang, Xie, Rosa, Guo, Wang, Trigoni, and
  Markham}]{hu2020randla}
Hu, Q., Yang, B., Xie, L., Rosa, S., Guo, Y., Wang, Z., Trigoni, N., Markham,
  A., 2020. Randla-net: Efficient semantic segmentation of large-scale point
  clouds. In: Proceedings of the IEEE Conference on Computer Vision and Pattern
  Recognition. pp. 11108--11117.

\bibitem[{{Huang} et~al.(2019){Huang}, {Hong}, {Xu}, {Yao}, and
  {Stilla}}]{huang2019embedding}
{Huang}, R., {Hong}, D., {Xu}, Y., {Yao}, W., {Stilla}, U., 2019. Multi-scale
  local context embedding for lidar point cloud classification. IEEE Geoscience
  and Remote Sensing Letters, 1--5.

\bibitem[{Huang et~al.(2020{\natexlab{a}})Huang, Xu, Hoegner, and
  Stilla}]{huang2020temporal}
Huang, R., Xu, Y., Hoegner, L., Stilla, U., 2020{\natexlab{a}}. Temporal
  comparison of construction sites using photogrammetric point cloud sequences
  and robust phase correlation. Automation in Construction 117, 103247.

\bibitem[{Huang et~al.(2020{\natexlab{b}})Huang, Xu, Hong, Yao, Ghamisi, and
  Stilla}]{huang2020deep}
Huang, R., Xu, Y., Hong, D., Yao, W., Ghamisi, P., Stilla, U.,
  2020{\natexlab{b}}. Deep point embedding for urban classification using als
  point clouds: A new perspective from local to global. ISPRS Journal of
  Photogrammetry and Remote Sensing 163, 62--81.

\bibitem[{Jiang et~al.(2018)Jiang, Wu, Zhao, Zhao, and Lu}]{jiang2018pointsift}
Jiang, M., Wu, Y., Zhao, T., Zhao, Z., Lu, C., 2018. Pointsift: A sift-like
  network module for 3d point cloud semantic segmentation. arXiv preprint
  arXiv:1807.00652.

\bibitem[{Jutzi and Gross(2010)}]{jutzi2010investigations}
Jutzi, B., Gross, H., 2010. Investigations on surface reflection models for
  intensity normalization in airborne laser scanning (als) data.
  Photogrammetric Engineering \& Remote Sensing 76~(9), 1051--1060.

\bibitem[{Klokov and Lempitsky(2017)}]{klokov2017escape}
Klokov, R., Lempitsky, V., 2017. Escape from cells: Deep kd-networks for the
  recognition of 3d point cloud models. In: Proceedings of the IEEE
  International Conference on Computer Vision. pp. 863--872.

\bibitem[{Lafarge and Mallet(2012)}]{lafarge2012creating}
Lafarge, F., Mallet, C., 2012. Creating large-scale city models from 3d-point
  clouds: a robust approach with hybrid representation. International Journal
  of Computer Vision 99~(1), 69--85.

\bibitem[{Landrieu et~al.(2017)Landrieu, Raguet, Vallet, Mallet, and
  Weinmann}]{landrieu2017structured}
Landrieu, L., Raguet, H., Vallet, B., Mallet, C., Weinmann, M., 2017. A
  structured regularization framework for spatially smoothing semantic
  labelings of 3d point clouds. ISPRS Journal of Photogrammetry and Remote
  Sensing 132, 102--118.

\bibitem[{Landrieu and Simonovsky(2018)}]{landrieu2018large}
Landrieu, L., Simonovsky, M., 2018. Large-scale point cloud semantic
  segmentation with superpoint graphs. In: Proceedings of the IEEE Conference
  on Computer Vision and Pattern Recognition. pp. 4558--4567.

\bibitem[{Li et~al.(2019{\natexlab{a}})Li, Liu, and Pfeifer}]{li2019improving}
Li, N., Liu, C., Pfeifer, N., 2019{\natexlab{a}}. Improving lidar
  classification accuracy by contextual label smoothing in post-processing.
  ISPRS Journal of Photogrammetry and Remote Sensing 148, 13--31.

\bibitem[{Li et~al.(2020{\natexlab{a}})Li, Wang, and Xia}]{li2020geometry}
Li, W., Wang, F.-D., Xia, G.-S., 2020{\natexlab{a}}. A geometry-attentional
  network for als point cloud classification. ISPRS Journal of Photogrammetry
  and Remote Sensing 164, 26--40.

\bibitem[{Li et~al.(2013)Li, Cheng, Liu, Xiao, Ma, Jin, Che, Liu, Wang, Qi,
  et~al.}]{li2013heihe}
Li, X., Cheng, G., Liu, S., Xiao, Q., Ma, M., Jin, R., Che, T., Liu, Q., Wang,
  W., Qi, Y., et~al., 2013. Heihe watershed allied telemetry experimental
  research (hiwater): Scientific objectives and experimental design. Bulletin
  of the American Meteorological Society 94~(8), 1145--1160.

\bibitem[{Li et~al.(2020{\natexlab{b}})Li, Wang, Wang, Wen, and
  Fang}]{li2020dance}
Li, X., Wang, L., Wang, M., Wen, C., Fang, Y., 2020{\natexlab{b}}. Dance-net:
  Density-aware convolution networks with context encoding for airborne lidar
  point cloud classification. ISPRS Journal of Photogrammetry and Remote
  Sensing 166, 128--139.

\bibitem[{Li et~al.(2018)Li, Bu, Sun, Wu, Di, and Chen}]{li2018pointcnn}
Li, Y., Bu, R., Sun, M., Wu, W., Di, X., Chen, B., 2018. Pointcnn: Convolution
  on x-transformed points. In: Advances in Neural Information Processing
  Systems. pp. 820--830.

\bibitem[{Li et~al.(2019{\natexlab{b}})Li, Chen, Du, Xia, Wang, Xu, and
  Yang}]{li2019higher}
Li, Y., Chen, D., Du, X., Xia, S., Wang, Y., Xu, S., Yang, Q.,
  2019{\natexlab{b}}. Higher-order conditional random fields-based 3d semantic
  labeling of airborne laser-scanning point clouds. Remote Sensing 11~(10),
  1248.

\bibitem[{Maas(1999)}]{maas1999potential}
Maas, H.-G., 1999. The potential of height texture measures for the
  segmentation of airborne laserscanner data. In: Fourth International Airborne
  Remote Sensing Conference and Exhibition/21st Canadian Symposium on Remote
  Sensing. Vol.~1. pp. 154--161.

\bibitem[{Mallet et~al.(2011)Mallet, Bretar, Roux, Soergel, and
  Heipke}]{mallet2011relevance}
Mallet, C., Bretar, F., Roux, M., Soergel, U., Heipke, C., 2011. Relevance
  assessment of full-waveform lidar data for urban area classification. ISPRS
  Journal of Photogrammetry and Remote Sensing 66~(6), S71--S84.

\bibitem[{Maturana and Scherer(2015)}]{maturana2015voxnet}
Maturana, D., Scherer, S., 2015. Voxnet: A 3d convolutional neural network for
  real-time object recognition. In: 2015 IEEE/RSJ International Conference on
  Intelligent Robots and Systems (IROS). IEEE, pp. 922--928.

\bibitem[{Moussa and El-Sheimy(2010)}]{moussa2010automatic}
Moussa, A.~M., El-Sheimy, N., 2010. Automatic classification and 3d modeling of
  lidar data. In: Proceedings of the ISPRS Commission III symposium. Vol.~38.
  pp. 155--159.

\bibitem[{Niemeyer et~al.(2014)Niemeyer, Rottensteiner, and
  Soergel}]{niemeyer2014contextual}
Niemeyer, J., Rottensteiner, F., Soergel, U., 2014. Contextual classification
  of lidar data and building object detection in urban areas. ISPRS Journal of
  Photogrammetry and Remote Sensing 87, 152--165.

\bibitem[{Niemeyer et~al.(2016)Niemeyer, Rottensteiner, S{\"o}rgel, and
  Heipke}]{niemeyer2016hierarchical}
Niemeyer, J., Rottensteiner, F., S{\"o}rgel, U., Heipke, C., 2016. Hierarchical
  higher order crf for the classification of airborne lidar point clouds in
  urban areas. International Archives of Photogrammetry, Remote Sensing and
  Spatial Information Sciences XLI-B3, 655--662.

\bibitem[{Olsen et~al.(2010)Olsen, Kuester, Chang, and
  Hutchinson}]{olsen2010terrestrial}
Olsen, M.~J., Kuester, F., Chang, B.~J., Hutchinson, T.~C., 2010. Terrestrial
  laser scanning-based structural damage assessment. Journal of Computing in
  Civil Engineering 24~(3), 264--272.

\bibitem[{Polewski et~al.(2015)Polewski, Yao, Heurich, Krzystek, and
  Stilla}]{polewski2015detection}
Polewski, P., Yao, W., Heurich, M., Krzystek, P., Stilla, U., 2015. Detection
  of fallen trees in als point clouds using a normalized cut approach trained
  by simulation. ISPRS Journal of Photogrammetry and Remote Sensing 105,
  252--271.

\bibitem[{Qi et~al.(2018)Qi, Liu, Wu, Su, and Guibas}]{qi2018frustum}
Qi, C.~R., Liu, W., Wu, C., Su, H., Guibas, L.~J., 2018. Frustum pointnets for
  3d object detection from rgb-d data. In: Proceedings of the IEEE Conference
  on Computer Vision and Pattern Recognition. pp. 918--927.

\bibitem[{Qi et~al.(2017{\natexlab{a}})Qi, Su, Mo, and Guibas}]{qi2017pointnet}
Qi, C.~R., Su, H., Mo, K., Guibas, L.~J., 2017{\natexlab{a}}. Pointnet: Deep
  learning on point sets for 3d classification and segmentation. In:
  Proceedings of the IEEE Conference on Computer Vision and Pattern
  Recognition. pp. 652--660.

\bibitem[{Qi et~al.(2016)Qi, Su, Nie{\ss}ner, Dai, Yan, and
  Guibas}]{qi2016volumetric}
Qi, C.~R., Su, H., Nie{\ss}ner, M., Dai, A., Yan, M., Guibas, L.~J., 2016.
  Volumetric and multi-view cnns for object classification on 3d data. In:
  Proceedings of the IEEE Conference on Computer Vision and Pattern
  Recognition. pp. 5648--5656.

\bibitem[{Qi et~al.(2017{\natexlab{b}})Qi, Yi, Su, and
  Guibas}]{qi2017pointnet++}
Qi, C.~R., Yi, L., Su, H., Guibas, L.~J., 2017{\natexlab{b}}. Pointnet++: Deep
  hierarchical feature learning on point sets in a metric space. In: Advances
  in Neural Information Processing Systems. pp. 5099--5108.

\bibitem[{Qin et~al.(2019)Qin, Hu, Wang, Shan, and Li}]{qin2019semantic}
Qin, N., Hu, X., Wang, P., Shan, J., Li, Y., 2019. Semantic labeling of als
  point cloud via learning voxel and pixel representations. IEEE Geoscience and
  Remote Sensing Letters.

\bibitem[{Rabbani et~al.(2006)Rabbani, Van Den~Heuvel, and
  Vosselmann}]{rabbani2006segmentation}
Rabbani, T., Van Den~Heuvel, F., Vosselmann, G., 2006. Segmentation of point
  clouds using smoothness constraint. International archives of photogrammetry,
  remote sensing and spatial information sciences 36~(5), 248--253.

\bibitem[{Reitberger et~al.(2009)Reitberger, Schn{\"o}rr, Krzystek, and
  Stilla}]{reitberger20093d}
Reitberger, J., Schn{\"o}rr, C., Krzystek, P., Stilla, U., 2009. 3d
  segmentation of single trees exploiting full waveform lidar data. ISPRS
  Journal of Photogrammetry and Remote Sensing 64~(6), 561--574.

\bibitem[{Rottensteiner et~al.(2012)Rottensteiner, Sohn, Jung, Gerke, Baillard,
  Benitez, and Breitkopf}]{rottensteiner2012isprs}
Rottensteiner, F., Sohn, G., Jung, J., Gerke, M., Baillard, C., Benitez, S.,
  Breitkopf, U., 2012. The isprs benchmark on urban object classification and
  3d building reconstruction. ISPRS Annals of the Photogrammetry, Remote
  Sensing and Spatial Information Sciences I-3 (2012), Nr. 1, 293--298.

\bibitem[{Simonovsky and Komodakis(2017)}]{simonovsky2017dynamic}
Simonovsky, M., Komodakis, N., 2017. Dynamic edge-conditioned filters in
  convolutional neural networks on graphs. In: Proceedings of the IEEE
  Conference on Computer Vision and Pattern Recognition. pp. 3693--3702.

\bibitem[{Su et~al.(2015)Su, Maji, Kalogerakis, and
  Learned-Miller}]{su2015multi}
Su, H., Maji, S., Kalogerakis, E., Learned-Miller, E., 2015. Multi-view
  convolutional neural networks for 3d shape recognition. In: Proceedings of
  the IEEE international conference on computer vision. pp. 945--953.

\bibitem[{Sun et~al.(2018)Sun, Xu, Hoegner, and Stilla}]{sun2018classification}
Sun, Z., Xu, Y., Hoegner, L., Stilla, U., 2018. Classification of mls point
  cloud in urban scenes using detrended geometric features from
  supervoxel-based local contexts. ISPRS Annals of Photogrammetry, Remote
  Sensing and Spatial Information Sciences 4~(2), 271--278.

\bibitem[{Vosselman et~al.(2017)Vosselman, Coenen, and
  Rottensteiner}]{vosselman2017contextual}
Vosselman, G., Coenen, M., Rottensteiner, F., 2017. Contextual segment-based
  classification of airborne laser scanner data. ISPRS Journal of
  Photogrammetry and Remote Sensing 128, 354--371.

\bibitem[{Wang and Posner(2015)}]{wang2015voting}
Wang, D.~Z., Posner, I., July 2015. Voting for voting in online point cloud
  object detection. In: Proceedings of Robotics: Science and Systems. Rome,
  Italy.

\bibitem[{Wang et~al.(2017{\natexlab{a}})Wang, Jiang, Qian, Yang, Li, Zhang,
  Wang, and Tang}]{wang2017residual}
Wang, F., Jiang, M., Qian, C., Yang, S., Li, C., Zhang, H., Wang, X., Tang, X.,
  2017{\natexlab{a}}. Residual attention network for image classification. In:
  Proceedings of the IEEE conference on computer vision and pattern
  recognition. pp. 3156--3164.

\bibitem[{Wang et~al.(2017{\natexlab{b}})Wang, Liu, Guo, Sun, and
  Tong}]{wang2017cnn}
Wang, P.-S., Liu, Y., Guo, Y.-X., Sun, C.-Y., Tong, X., 2017{\natexlab{b}}.
  O-cnn: Octree-based convolutional neural networks for 3d shape analysis. ACM
  Transactions on Graphics 36~(4), 72.

\bibitem[{Wang et~al.(2019)Wang, Sun, Liu, Sarma, Bronstein, and
  Solomon}]{wang2018dynamic}
Wang, Y., Sun, Y., Liu, Z., Sarma, S.~E., Bronstein, M.~M., Solomon, J.~M.,
  2019. Dynamic graph cnn for learning on point clouds. ACM Transactions on
  Graphics 38~(5), 1--12.

\bibitem[{Weinmann et~al.(2015{\natexlab{a}})Weinmann, Jutzi, Hinz, and
  Mallet}]{weinmann2015semantic}
Weinmann, M., Jutzi, B., Hinz, S., Mallet, C., 2015{\natexlab{a}}. Semantic
  point cloud interpretation based on optimal neighborhoods, relevant features
  and efficient classifiers. ISPRS Journal of Photogrammetry and Remote Sensing
  105, 286--304.

\bibitem[{Weinmann et~al.(2015{\natexlab{b}})Weinmann, Schmidt, Mallet, Hinz,
  Rottensteiner, and Jutzi}]{weinmann2015contextual}
Weinmann, M., Schmidt, A., Mallet, C., Hinz, S., Rottensteiner, F., Jutzi, B.,
  2015{\natexlab{b}}. Contextual classification of point cloud data by
  exploiting individual 3d neigbourhoods. ISPRS Annals of the Photogrammetry,
  Remote Sensing and Spatial Information Sciences II-3 (2015), Nr. W4 2~(W4),
  271--278.

\bibitem[{Weinmann et~al.(2015{\natexlab{c}})Weinmann, Urban, Hinz, Jutzi, and
  Mallet}]{weinmann2015distinctive}
Weinmann, M., Urban, S., Hinz, S., Jutzi, B., Mallet, C., 2015{\natexlab{c}}.
  Distinctive 2d and 3d features for automated large-scale scene analysis in
  urban areas. Computers \& Graphics 49, 47--57.

\bibitem[{Wen et~al.(2020)Wen, Yang, Li, Peng, and Chi}]{wen2020directionally}
Wen, C., Yang, L., Li, X., Peng, L., Chi, T., 2020. Directionally constrained
  fully convolutional neural network for airborne lidar point cloud
  classification. ISPRS Journal of Photogrammetry and Remote Sensing 162,
  50--62.

\bibitem[{Xu et~al.(2019{\natexlab{a}})Xu, Jiang, Liang, and
  Li}]{xu2019spatial}
Xu, H., Jiang, C., Liang, X., Li, Z., 2019{\natexlab{a}}. Spatial-aware graph
  relation network for large-scale object detection. In: Proceedings of the
  IEEE Conference on Computer Vision and Pattern Recognition. pp. 9298--9307.

\bibitem[{Xu et~al.(2018)Xu, Hoegner, Tuttas, and Stilla}]{xu2018a}
Xu, Y., Hoegner, L., Tuttas, S., Stilla, U., 2018. A voxel-and graph-based
  strategy for segmenting man-made infrastructures using perceptual grouping
  laws: Comparison and evaluation. Photogrammetric Engineering \& Remote
  Sensing 84~(6), 377--391.

\bibitem[{Xu et~al.(2019{\natexlab{b}})Xu, Ye, Yao, Huang, Tong, Hoegner, and
  Stilla}]{xu2019classification}
Xu, Y., Ye, Z., Yao, W., Huang, R., Tong, X., Hoegner, L., Stilla, U.,
  2019{\natexlab{b}}. Classification of lidar point clouds using
  supervoxel-based detrended feature and perception-weighted graphical model.
  IEEE Journal of Selected Topics in Applied Earth Observations and Remote
  Sensing~(99), 1--17.

\bibitem[{Yan et~al.(2015)Yan, Shaker, and El-Ashmawy}]{yan2015urban}
Yan, W.~Y., Shaker, A., El-Ashmawy, N., 2015. Urban land cover classification
  using airborne lidar data: A review. Remote Sensing of Environment 158,
  295--310.

\bibitem[{Yang et~al.(2017{\natexlab{a}})Yang, Dong, Liu, Liang, and
  Wang}]{yang2017computing}
Yang, B., Dong, Z., Liu, Y., Liang, F., Wang, Y., 2017{\natexlab{a}}. Computing
  multiple aggregation levels and contextual features for road facilities
  recognition using mobile laser scanning data. ISPRS Journal of Photogrammetry
  and Remote Sensing 126, 180--194.

\bibitem[{Yang et~al.(2013)Yang, Xu, and Dong}]{yang2013automated}
Yang, B., Xu, W., Dong, Z., 2013. Automated extraction of building outlines
  from airborne laser scanning point clouds. IEEE Geoscience and Remote Sensing
  Letters 10~(6), 1399--1403.

\bibitem[{Yang et~al.(2017{\natexlab{b}})Yang, Jiang, Xu, Zhu, Jiang, and
  Huang}]{yang2017convolutional}
Yang, Z., Jiang, W., Xu, B., Zhu, Q., Jiang, S., Huang, W., 2017{\natexlab{b}}.
  A convolutional neural network-based 3d semantic labeling method for als
  point clouds. Remote Sensing 9~(9), 936.

\bibitem[{Yang et~al.(2018)Yang, Tan, Pei, and Jiang}]{yang2018segmentation}
Yang, Z., Tan, B., Pei, H., Jiang, W., 2018. Segmentation and multi-scale
  convolutional neural network-based classification of airborne laser scanner
  data. Sensors 18~(10), 3347.

\bibitem[{Yao et~al.(2017)Yao, Polewski, and Krzystek}]{yao2017semantic}
Yao, W., Polewski, P., Krzystek, P., 2017. Semantic labeling of ultra dense mls
  point clouds in urban road corridors based on fusing crf with shape priors.
  International Archives of the Photogrammetry, Remote Sensing and Spatial
  Information Sciences 42, 971--976.

\bibitem[{Ye et~al.(2020)Ye, Xu, Huang, Tong, Li, Liu, Luan, Hoegner, and
  Stilla}]{ye2020large}
Ye, Z., Xu, Y., Huang, R., Tong, X., Li, X., Liu, X., Luan, K., Hoegner, L.,
  Stilla, U., 2020. Lasdu: A large-scale aerial lidar dataset for semantic
  labeling in dense urban areas. ISPRS International Journal of Geo-Information
  9~(7).
\newline\urlprefix\url{https://www.mdpi.com/2220-9964/9/7/450}

\bibitem[{Yousefhussien et~al.(2018)Yousefhussien, Kelbe, Ientilucci, and
  Salvaggio}]{yousefhussien2018multi}
Yousefhussien, M., Kelbe, D.~J., Ientilucci, E.~J., Salvaggio, C., 2018. A
  multi-scale fully convolutional network for semantic labeling of 3d point
  clouds. ISPRS Journal of Photogrammetry and Remote Sensing 143, 191--204.

\bibitem[{Yu et~al.(2016)Yu, Li, Wen, Guan, Luo, and Wang}]{yu2016bag}
Yu, Y., Li, J., Wen, C., Guan, H., Luo, H., Wang, C., 2016.
  Bag-of-visual-phrases and hierarchical deep models for traffic sign detection
  and recognition in mobile laser scanning data. ISPRS Journal of
  Photogrammetry and Remote Sensing 113, 106--123.

\bibitem[{Zhang et~al.(2011)Zhang, de~Gier, Xing, and Sohn}]{zhang2011full}
Zhang, J., de~Gier, A., Xing, Y., Sohn, G., 2011. Full waveform-based analysis
  for forest type information derivation from large footprint spaceborne lidar
  data. Photogrammetric Engineering \& Remote Sensing 77~(3), 281--290.

\bibitem[{Zhang et~al.(2020)Zhang, Lan, Zeng, Jin, and
  Chen}]{zhang2020relation}
Zhang, Z., Lan, C., Zeng, W., Jin, X., Chen, Z., 2020. Relation-aware global
  attention for person re-identification. In: Proceedings of the IEEE/CVF
  Conference on Computer Vision and Pattern Recognition. pp. 3186--3195.

\bibitem[{Zhang et~al.(2019)Zhang, Sun, Zhong, Chen, Xu, Wang, Qin, Sun, and
  Li}]{zhang20193}
Zhang, Z., Sun, L., Zhong, R., Chen, D., Xu, Z., Wang, C., Qin, C.-Z., Sun, H.,
  Li, R., 2019. 3-d deep feature construction for mobile laser scanning point
  cloud registration. IEEE Geoscience and Remote Sensing Letters 16~(12),
  1904--1908.

\bibitem[{Zhao et~al.(2018)Zhao, Pang, and Wang}]{zhao2018classifying}
Zhao, R., Pang, M., Wang, J., 2018. Classifying airborne lidar point clouds via
  deep features learned by a multi-scale convolutional neural network.
  International Journal of Geographical Information Science 32~(5), 960--979.

\bibitem[{Zhou and Tuzel(2018)}]{zhou2018voxelnet}
Zhou, Y., Tuzel, O., 2018. Voxelnet: End-to-end learning for point cloud based
  3d object detection. In: Proceedings of the IEEE Conference on Computer
  Vision and Pattern Recognition. pp. 4490--4499.

\end{thebibliography}
\end{document}